\documentclass[dvipsnames]{article}
\usepackage{colm2024_conference}
\usepackage[T1]{fontenc} % 参考文献作者名中的 \k{a} (ą) 等字符在默认 OT1 编码下不可用
\pdfoutput=1

\usepackage{microtype}
\usepackage{graphicx}
\usepackage{subcaption}
\usepackage{booktabs} % for professional tables

\usepackage{placeins}  % 模板的 \mymaketitle 和下面的 \section 重定义用到 \FloatBarrier
\usepackage{tocloft}

\usepackage{amsmath}
\usepackage{amssymb}
\usepackage{mathtools}
\usepackage{amsthm}
\usepackage{multirow}

\usepackage[capitalize,noabbrev]{cleveref}

\theoremstyle{plain}
\newtheorem{theorem}{Theorem}[section]
\newtheorem{proposition}[theorem]{Proposition}

\theoremstyle{definition}

\theoremstyle{remark}

\crefname{hypothesis}{Hypothesis}{Hypotheses}
\crefname{proposition}{Proposition}{Propositions}

\usepackage{array} % p{} columns with \raggedright in tables
\usepackage{xspace}
\usepackage[most]{tcolorbox}
\usepackage{enumitem}
 \usepackage{colortbl}

\usepackage[textsize=tiny]{todonotes}

\definecolor{opdcol}{HTML}{1F5FA9}   % OPD (blue)
\definecolor{rlcol}{HTML}{A8431F}    % RL  (rust)
\definecolor{best}{HTML}{2A78D6}     % best cell shading: \cellcolor{best!18}
\definecolor{accent}{HTML}{E34948}   % failure cases / warnings
\definecolor{ink}{HTML}{52514E}      % frames, axis-grey text
\definecolor{midgrey}{HTML}{898781}  % baselines
\definecolor{gridgrey}{HTML}{C3C2B7}
\definecolor{paper}{HTML}{E1E0D9}    % light fill
\definecolor{opdlow}{HTML}{9DBBE0}\definecolor{opdmed}{HTML}{5B8CC8}\definecolor{opdhigh}{HTML}{1F5FA9}
\definecolor{rllow}{HTML}{E3A98F}\definecolor{rlmed}{HTML}{C9714C}\definecolor{rlhigh}{HTML}{A8431F}
\definecolor{phenobg}{HTML}{EBF5FB}\definecolor{phenoframe}{HTML}{1F5FA9}
\definecolor{mechbg}{HTML}{FEF5E7}\definecolor{mechframe}{HTML}{A8431F}
\definecolor{recipebg}{HTML}{EAFAF1}\definecolor{recipeframe}{HTML}{1E8449}
\definecolor{lightblue}{RGB}{220,235,250}
\definecolor{questionbg}{HTML}{F3EDF9}\definecolor{questionframe}{HTML}{5B3A8C}

\newcommand{\win}[1]{\cellcolor{best!18}\bfseries #1}
\newcommand{\lvl}[1]{\textsf{#1}}                % effort tier names
\newcommand{\low}{\lvl{low}\xspace}
\newcommand{\med}{\lvl{medium}\xspace}
\newcommand{\high}{\lvl{high}\xspace}

\newcommand{\E}{\mathbb{E}}

\newcommand{\KL}{D_{\mathrm{KL}}}

\newcommand{\dB}{d_{\mathrm{B}}}            % Bernoulli KL
\newcommand{\eot}{\texttt{</think>}\xspace} % end-of-thinking token

\newcommand{\pis}{\pi_\theta}
\newcommand{\Lth}{L^{\mathrm{th}}}          % thinking length

\tcbset{
  rqbox/.style={
    enhanced, colback=white, colframe=gridgrey, boxrule=0.5pt, arc=2mm,
    fontupper=\small, left=2mm, right=2mm, top=1mm, bottom=1mm},
  hypbox/.style={
    enhanced, colback=white, colframe=midgrey, boxrule=0.4pt, arc=1.5mm,
    borderline west={2pt}{0pt}{ink}, fontupper=\small,
    left=2mm, right=2mm, top=0.8mm, bottom=0.8mm},
  takeawaysbox/.style={
    title=Takeaways,
    colback=lightblue!80,
    colframe=black,
    fonttitle=\bfseries\small,
    coltitle=white,
    colbacktitle=black,
    enhanced,
    attach boxed title to top left={xshift=2.5mm,yshift=-2.5mm},
    boxed title style={rounded corners, size=small, colframe=black, colback=black},
    width=\linewidth,
    arc=3.5mm
  },
  questionsbox/.style={
    title=Questions,
    colback=questionbg,
    colframe=questionframe,
    fonttitle=\bfseries\small,
    coltitle=white,
    colbacktitle=questionframe,
    enhanced,
    attach boxed title to top left={xshift=2.5mm,yshift=-2.5mm},
    boxed title style={rounded corners, size=small, colframe=questionframe, colback=questionframe},
    width=\linewidth,
    arc=3.5mm
  }
}

\title{\begin{tabular}[t]{@{}c@{}}Rethinking Reasoning Effort Training\\in Large Language Models:\\On-Policy Distillation or Reinforcement Learning?\end{tabular}}

\author{%
\makebox[\linewidth][c]{%
\begin{tabular}{@{}c@{}}
\bf Tianliang Lv$^{*}$, Hao Li$^{*}$, Xinchao Xu, Zhengkun Zhang, Ju Huo$^{\dagger}$ \\[0.3em]
\textbf{ERNIE Team, Baidu}
\end{tabular}}%
}

\usepackage{etoolbox}
\newcommand{\redate}{Nov 24, 2026}
\newcommand{\redatefull}{November 24, 2026}
\newcommand{\remodelurl}{\url{https://huggingface.co/baidu/ERNIE-5.5-Flash-Preview}}

\patchcmd{\mymaketitle}{\small \textbf{Feb 4, 2026}}{%
  \small\raggedright
  \textbf{Date:} \redatefull\par\vspace{0.3em}
  \textbf{Model:} \remodelurl\par\vspace{0.3em}
}{}{\PackageWarning{re}{date patch failed}}
\makeatletter
\fancypagestyle{normalpage}{
  \fancyhf{}
  \lhead{\includegraphics[height=15pt]{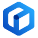}}
  \chead{\textcolor[HTML]{4D4D4D}{\textbf{\thetitle}}}
  \rhead{\textcolor[HTML]{4D4D4D}{\textbf{\redate}}}
  \renewcommand{\headrulewidth}{0.4pt}
  \renewcommand{\headrule}{{\color[HTML]{4D4D4D}\hrule\@height\headrulewidth\@width\headwidth\vskip-\headrulewidth}}
  \cfoot{\thepage}
}
\makeatother

\definecolor{refSec}{HTML}{9B1B1B}   % sections (dark red)
\definecolor{refApp}{HTML}{9B1B1B}   % appendix (dark red, same as sections)
\definecolor{refFig}{HTML}{7B3FA0}   % figures (purple)
\definecolor{refTab}{HTML}{7B3FA0}   % tables (purple, same as figures)
\definecolor{refEq}{HTML}{5A5A5A}    % equations and propositions
\definecolor{refCite}{HTML}{2E48C0}  % citations (slightly darker than wenxinblue 3D58DB)
\hypersetup{citecolor=refCite}
\makeatletter
\ifdefined\pdfmatch
\protected\def\re@pickcolor#1#2{%
  \def\re@type{#1}\def\re@link{link}%
  \ifx\re@type\re@link
    \ifnum\pdfmatch{^(appendix[.]|(sub)*section[.][A-Z])}{#2}=1 \def\@linkcolor{refApp}%
    \else\ifnum\pdfmatch{^(sub)*section[.]}{#2}=1 \def\@linkcolor{refSec}%
    \else\ifnum\pdfmatch{^figure[.]}{#2}=1 \def\@linkcolor{refFig}%
    \else\ifnum\pdfmatch{^table[.]}{#2}=1 \def\@linkcolor{refTab}%
    \else\ifnum\pdfmatch{^(equation|theorem|AMS)[.]}{#2}=1 \def\@linkcolor{refEq}%
    \else \def\@linkcolor{refSec}%
    \fi\fi\fi\fi\fi
  \fi}
\let\re@orig@linkstart\hyper@linkstart
\protected\def\hyper@linkstart#1#2{\re@pickcolor{#1}{#2}\re@orig@linkstart{#1}{#2}}
\let\re@orig@link\hyper@link
\protected\def\hyper@link#1#2{\re@pickcolor{#1}{#2}\re@orig@link{#1}{#2}}
\fi
\makeatother

\makeatletter\def\@abstract{
Large language models have shown strong capabilities on complex tasks, yet how to control their reasoning effort remains a significant challenge for practical deployment. This paper provides a systematic investigation of reasoning-effort training, contrasting reinforcement learning (RL) with on-policy distillation (OPD). We find that compressing reasoning and separating effort tiers are governed by different signals. Length-penalized RL reliably compresses a single policy, reducing thinking on AIME25 by 70\% without accuracy loss, but collapses the tiers when they are trained jointly. Multi-teacher OPD separates them: across 33 benchmarks, low and medium use 58\% and 43\% fewer thinking tokens than high, while high improves on the reference policy by 3.1 points. We trace this difference to the training signal. A sequence-level reward cannot tell tier-specific behavior from shared behavior, whereas OPD supplies a tier-specific target at every visited state, including the decision to stop. Building on this analysis, we show that a single token-level KL anchor on the high tier is sufficient to learn separated tiers within one RL stage, and that natural-language effort instructions make this separation emerge from the start of training, yielding a policy competitive with multi-teacher OPD at matched thinking length. Finally, the separation built by dense targets holds only on the states seen in training: it does not carry over to unseen agentic tasks, where effort acts mainly through the number of interaction turns, whereas the compression induced by sequence-level rewards does.

}\makeatother

\let\reOldSection\section
\renewcommand{\section}{\FloatBarrier\reOldSection}
\let\reOldAppendix\appendix
\renewcommand{\appendix}{%
  \reOldAppendix
  \crefname{section}{Appendix}{Appendices}%
  \Crefname{section}{Appendix}{Appendices}%
  \crefname{subsection}{Appendix}{Appendices}%
  \Crefname{subsection}{Appendix}{Appendices}%
  \crefname{subsubsection}{Appendix}{Appendices}%
  \Crefname{subsubsection}{Appendix}{Appendices}%
}

\begin{document}
\begingroup
\renewcommand\thefootnote{}
\footnotetext{%
  $^{*}$Equal Contribution. \quad %$^{\dagger}$Project Lead. \quad 
  $^{\dagger}$Corresponding Author.\newline
  \hspace*{2em}Contact email: 
  \href{mailto:tianl@alumni.pku.edu.cn}{tianl@alumni.pku.edu.cn}, 
  \href{mailto:h.li1@imperial.ac.uk}{h.li1@imperial.ac.uk} or 
  \href{mailto:haoli47@outlook.com}{haoli47@outlook.com}
}
\endgroup
% the template's \mymaketitle calls \tcbset globally (frame hidden, wenxinbg, 1cm padding);
% the group keeps those settings from leaking into the Takeaways boxes
\begingroup\maketitle\endgroup
% the full title overruns the date in the running header
\renewcommand{\thetitle}{Rethinking Reasoning Effort Training in LLMs}

\pagestyle{normalpage}
% Figure 1 sits on the title page, directly under the abstract box
  % \begin{figure}[!h]
  % \centering
  % \includegraphics[width=\linewidth]{figs/hero_findings.pdf}
  % \caption{\textbf{Summary of core findings.} (a)~Length dynamics (\cref{sec:phenomenology}); (b)~cross-tier KL at reflection and stopping tokens (\cref{sec:mechanism}); (c)~single-stage separation with a dense anchor (\cref{sec:dense-rl}).}
  % \label{fig:hero}
  % \end{figure}

\clearpage
\pagestyle{firstpage}
\begingroup
\hypersetup{hidelinks,linktoc=all}
\color{wenxinblue}
\renewcommand{\cftsecfont}{\bfseries\color{wenxinblue}}
\renewcommand{\cftsecpagefont}{\cftsecfont}
\renewcommand{\cftsubsecfont}{\color{wenxinblue}}
\renewcommand{\cftsubsecpagefont}{\cftsubsecfont}
\renewcommand{\cftsubsubsecfont}{\color{wenxinblue}}
\renewcommand{\cftsubsubsecpagefont}{\cftsubsubsecfont}
\tableofcontents
\endgroup
\clearpage
\pagestyle{normalpage}

\section{Introduction}
\label{sec:intro}

Scaling test-time compute through extended chains of thought has improved the performance of large reasoning models on challenging tasks~\citep{openai2024gpt4ocard, DBLP:journals/nature/GuoYZSWZXZMBZY025,DBLP:journals/corr/abs-2505-09388}. These gains come with increased inference latency and token usage, and additional reasoning does not always improve the answer. Models may repeatedly verify established conclusions or continue reasoning after sufficient evidence has been obtained~\citep{DBLP:journals/corr/abs-2412-21187,DBLP:journals/tmlr/SuiCWZZYLWZZCH25}. This motivates controlling reasoning effort according to task requirements~\citep{DBLP:journals/corr/abs-2510-27042}. Existing approaches include binary thinking controls~\citep{DBLP:conf/emnlp/ZhangLHFL25,DBLP:journals/corr/abs-2505-11896,DBLP:journals/corr/abs-2504-09858}, budget-conditioned generation~\citep{DBLP:conf/emnlp/MuennighoffYSLFHZLCH25,DBLP:journals/corr/abs-2503-04697}, and discrete effort-level conditioning~\citep{DBLP:journals/corr/abs-2508-10925,DBLP:journals/corr/abs-2508-18773}. Their common objective is a single policy that supports multiple accuracy--compute operating points.

Single-policy compression is typically achieved through supervised fine-tuning on shortened reasoning traces~\citep{DBLP:conf/emnlp/XiaLWLL25,DBLP:conf/acl/MaWYFW25} or reinforcement learning (RL) with length-dependent penalties~\citep{DBLP:journals/corr/abs-2501-12599,DBLP:conf/nips/AroraZ25,DBLP:journals/tmlr/HouZJLQAC26,DBLP:conf/acl/YuanXHZGLWZ26,DBLP:journals/corr/abs-2601-06052,DBLP:journals/corr/abs-2602-20945}. To obtain multiple effort tiers, Mixed RL conditions a shared policy on the requested tier and applies tier-specific rewards~\citep{DBLP:journals/corr/abs-2606-19348}. However, successful compression of individual policies does not ensure that their operating points can be learned jointly: updates across effort tiers can interfere, degrading the highest-effort tier~\citep{DBLP:journals/corr/abs-2508-18773}. An alternative is to train budget-specific teachers and consolidate them through multi-teacher on-policy distillation (OPD)~\citep{DBLP:journals/corr/abs-2601-08310}. Unlike trajectory-level RL rewards, OPD provides token-level supervision on student-generated states, extending a paradigm used to consolidate domain specialists in post-training~\citep{DBLP:journals/corr/abs-2607-24653,DBLP:journals/corr/abs-2601-02780}.

Despite this progress, compression and multi-tier control exhibit different optimization challenges. Prior work shows that RL can compress individual policies and transfer the resulting reductions to held-out domains~\citep{DBLP:journals/corr/abs-2601-06052}. Yet multi-budget RL can converge to nearly indistinguishable effort modes~\citep{DBLP:journals/corr/abs-2601-08310}, and inadequate mode-conditioned initialization can degrade high-effort performance~\citep{DBLP:journals/corr/abs-2508-18773}. Multi-teacher OPD offers a way to consolidate budget-specific behaviors~\citep{DBLP:journals/corr/abs-2601-08310}, but these results do not establish whether the resulting control transfers from single-turn reasoning to unseen multi-turn interaction formats. This leaves two questions.
\begin{tcolorbox}[
    enhanced, colback=black!2, colframe=black!25,
    boxrule=0.5pt, arc=1mm,
    left=3mm, right=3mm, top=2mm, bottom=2mm,
    before skip=8pt, after skip=8pt]
\begin{enumerate}[label=\textbf{RQ\arabic*.},leftmargin=*,topsep=0pt,itemsep=5pt,parsep=0pt]
\item \textbf{What prevents successful single-policy compression from yielding reliable multi-tier control?}
\item \textbf{Under what conditions does learned effort control generalize across interaction formats?}
\end{enumerate}
\end{tcolorbox}

\begin{figure}[!htb]
\centering
\includegraphics[width=\linewidth]{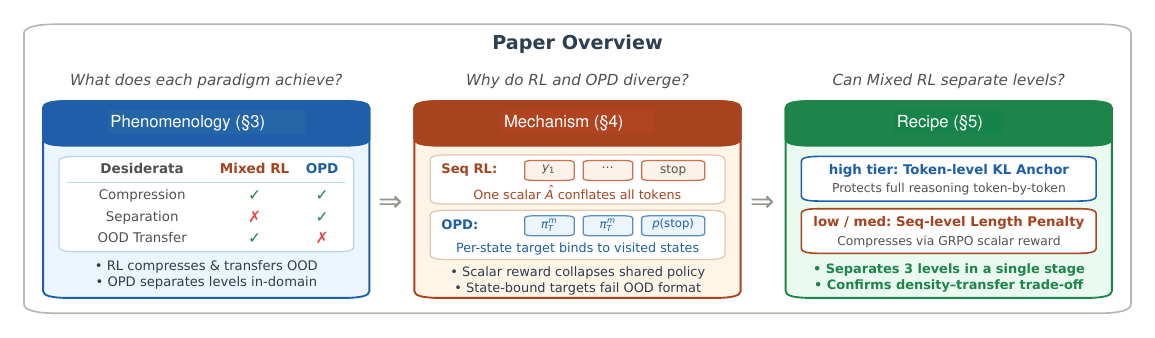}
\caption{\textbf{Overview of the study.} We compare sparse trajectory-level RL and dense token-level OPD across single-policy compression, multi-tier effort control, and transfer across interaction formats (\cref{sec:phenomenology}); analyze the reward and stopping constraints underlying these outcomes (\cref{sec:mechanism}); and develop dense-anchor RL recipes that preserve high-effort behavior while adapting lower-effort tiers (\cref{sec:dense-rl}).}
\label{fig:overview}
\end{figure}

To answer these questions, we present a systematic study of reasoning-effort training, progressing from empirical observations through objective-level analysis to practical training recipes (\cref{fig:overview}). Our contributions are threefold. First, we provide a controlled comparison of sparse trajectory-level RL and dense multi-teacher OPD for single-policy compression, multi-tier effort control, and transfer across interaction formats. Second, we identify the objective-level mechanism behind their different behaviors: sparse Mixed RL lacks a state-wise preservation signal for the high-effort tier, whereas dense OPD constrains stopping only on states represented during training. Third, we derive and evaluate two single-stage recipes that combine a dense high-tier anchor with sparse lower-tier compression, and further use natural-language effort instructions to accelerate tier separation and improve transfer. The findings and recipes of this paper support the reasoning-effort training of ERNIE-5.5-Flash-Preview\footnote{\url{https://huggingface.co/baidu/ERNIE-5.5-Flash-Preview}}.

% \begin{figure}[!htb]
% \centering
% \includegraphics[width=\linewidth]{figs/teaser.pdf}
% \caption{\textbf{Compression, controllability, and generalization are governed by supervisory signal density.} (a)~Mixed RL with sparse rewards across three effort tiers suffers from mode collapse to a uniform reasoning length (Qwen3-8B). (b)~Multi-teacher OPD with dense signals achieves clean multi-tier controllability; bar annotations indicate mean accuracy across 33 benchmarks. (c)~Median token ratio ($\rho_m$) of \low and \med relative to the highest effort tier, illustrating that sparse RL compression generalizes zero-shot to unseen agentic formats whereas dense controllability collapses\est{}.}
% \label{fig:teaser}
% \end{figure}

\textbf{Phenomenology (\cref{sec:phenomenology}).}
We identify a duality between compression and conditional effort control. (i)~Sparse length-penalized RL learns a global stopping prior: it shortens an individual policy while largely preserving task behavior, and the resulting compression transfers to held-out domains and unseen multi-turn agentic tasks. The same signal is insufficient for joint control, however. When lower and higher effort tiers share parameters, Mixed RL either collapses the tiers or leaves the lower tiers truncated and unstable. (ii)~Dense multi-teacher OPD can consolidate distinct operating points within one policy, but its controllability is tied to the interaction states represented during distillation. It separates effort tiers on held-out single-turn tasks yet can collapse or invert them on unseen agentic formats when those states are absent from training. This distinction separates two properties that are often conflated: the transfer of a compression prior and the transfer of conditional effort semantics.

\textbf{Mechanism (\cref{sec:mechanism}).}
We next identify the objective-level constraints that produce this duality. Sparse Mixed RL supplies shortening pressure to the lower tiers without an equivalent state-wise preservation constraint for \high. Group normalization can further remove the relative update when rollouts share both verifier outcome and length, while any remaining scalar advantage is broadcast across the entire trajectory. Shared parameters then allow lower-tier shortening updates to alter high-tier continuation behavior. Dense OPD supplies the missing local constraint: exact next-token KL bounds the divergence of the student's stopping probability from the teacher's on states that the student visits. This explains its sharper in-domain separation, but the guarantee is distributional rather than universal; it does not constrain stopping on unseen tool observations or interaction states. The resulting picture is a trade-off between supervisory density, which isolates conditional behavior, and state coverage, which determines where that behavior can transfer.

\textbf{Recipes (\cref{sec:dense-rl}).}
These observations suggest two complementary ways to make single-stage effort training effective. Recipe~I preserves the high-effort behavior with a dense token-level KL anchor while applying length-penalized RL to \low and \med. This addresses the asymmetric preservation problem, although discrete effort tags require substantial training before their meanings separate. Recipe~II makes those meanings explicit by pairing the same dense-anchor objective with natural-language descriptions of the intended reasoning depth. The resulting prompt-guided policies separate more quickly, achieve ordered effort tiers across the shared general and reasoning suite, and carry this ordering to multi-turn code and agent benchmarks although they are trained on single-turn data only. Together, the recipes turn the mechanistic diagnosis into a practical principle: preserve the high tier densely, adapt lower tiers with sparse length pressure, and specify effort semantics in the input when rapid control is required.

\section{Preliminaries}
\label{sec:prelim}

\subsection{Notation}
Let $x \sim \mathcal{D}$ denote a prompt and $y = (y_1, \dots, y_L)$ a response consisting of a reasoning trace terminated by \eot followed by a final answer, where $L \coloneqq |y|$ and $\Lth(y)$ denotes the thinking length preceding \eot.
We write $y_{<i} \coloneqq (y_1, \dots, y_{i-1})$ for the prefix before position $i$ and $s_i \coloneqq (x, y_{<i})$ for the corresponding state.
A policy $\pi$ defines a next-token distribution $\pi(\cdot \mid s_i)$ over a vocabulary $\mathcal{V}$, and $y \sim \pi(\cdot \mid x)$ denotes a response sampled autoregressively from $\pi$.
A verifier $v(x, y) \in \{0, 1\}$ indicates whether the final answer in $y$ is correct.
For two distributions $P$ and $Q$ over $\mathcal{V}$, the Kullback--Leibler (KL) divergence is $D_{\mathrm{KL}}(P \,\|\, Q) = \sum_{u \in \mathcal{V}} P(u) \log \frac{P(u)}{Q(u)}$.

\subsection{Reasoning Effort}
\label{sec:prelim-effort}

Let $\mathcal{M} = \{m_1, \dots, m_K\}$ be an ordered set of effort tiers with $m_1 \prec \dots \prec m_K$ (e.g., $\mathcal{M} = \{\low, \med, \high\}$), each associated with a reference token budget $B_m$ such that $B_{m_1} < \dots < B_{m_K}$.
An effort-controllable policy $\pi_\theta$ is conditioned on a requested tier $m \in \mathcal{M}$, and we write $\pi_\theta^m(\cdot \mid s_i) \coloneqq \pi_\theta(\cdot \mid s_i, m)$.
The goal is a single policy that, for every $m \in \mathcal{M}$, maximizes $\mathbb{E}_{x \sim \mathcal{D},\, y \sim \pi_\theta^m(\cdot \mid x)} [ v(x, y) ]$ while producing responses whose mean thinking length $\bar{L}_m \coloneqq \mathbb{E}[\Lth(y)]$ (and relative length ratio $\rho_m \coloneqq \bar{L}_m / \bar{L}_{\high}$) is commensurate with $B_m$.

\subsection{On-Policy Distillation}
\label{sec:prelim-opd}

On-policy distillation (OPD) trains a student $\pi_\theta$ toward a teacher $\pi_T$ using responses sampled from the student~\citep{DBLP:conf/iclr/AgarwalVZSGGB24}.
Given a prompt $x$ and a student response $y \sim \pi_\theta(\cdot \mid x)$, both models are evaluated on the student-generated states, yielding $p_i(\cdot) \coloneqq \pi_\theta(\cdot \mid s_i)$ and $q_i(\cdot) \coloneqq \pi_T(\cdot \mid s_i)$.
The standard objective is the sequence-level reverse KL divergence, which admits an exact token-level decomposition:
\begin{equation}
\begin{aligned}
\mathcal{L}_{\mathrm{OPD}}(\theta)
&= \mathbb{E}_{x \sim \mathcal{D}} \big[ D_{\mathrm{KL}}\big( \pi_\theta(\cdot \mid x) \,\|\, \pi_T(\cdot \mid x) \big) \big] \\
&= \mathbb{E}_{x \sim \mathcal{D},\, y \sim \pi_\theta(\cdot \mid x)} \Big[ \sum_{i=1}^{L} D_{\mathrm{KL}}( p_i \,\|\, q_i ) \Big].
\end{aligned}
\label{eq:opd}
\end{equation}
The most common implementation evaluates only the sampled token~\citep{DBLP:journals/air/FangYCCWLYLGLMRAZWLLZKZ26}, using
\begin{equation}
\ell_i = \log p_i(y_i) - \log q_i(y_i),
\label{eq:opd-sample}
\end{equation}
which satisfies $\mathbb{E}_{y_i \sim p_i}[\ell_i] = D_{\mathrm{KL}}(p_i \,\|\, q_i)$.
Equivalently, $A_i \coloneqq -\ell_i$ serves as a per-token advantage in a policy-gradient update, so that OPD is a special case of dense KL-constrained reinforcement learning with a teacher-induced token-level reward~\citep{DBLP:journals/corr/abs-2602-12125}.

\paragraph{Multi-teacher OPD.}
Given a set of teachers $\{\pi_T^c\}_{c \in \mathcal{C}}$ and a routing function $c(x) \in \mathcal{C}$, multi-teacher OPD replaces $q_i$ in \cref{eq:opd} with $q_i^{c(x)}(\cdot) \coloneqq \pi_T^{c(x)}(\cdot \mid s_i)$, thereby distilling all teachers into a single student~\citep{DBLP:journals/corr/abs-2606-30406}.
In existing pipelines, $c(x)$ typically indexes the domain of $x$; in our multi-tier effort setting, we instead route by the requested effort tier $m \in \mathcal{M}$, distilling tier-specific teachers $\{\pi_T^m\}_{m \in \mathcal{M}}$ into $\pi_\theta^m$.

\subsection{Reinforcement Learning with Verifiable Rewards}
\label{sec:prelim-rl}

Reinforcement learning with verifiable rewards (RLVR) maximizes the expected outcome reward of student responses:
\begin{equation}
\mathcal{J}_{\mathrm{RL}}(\theta) = \mathbb{E}_{x \sim \mathcal{D},\, y \sim \pi_\theta(\cdot \mid x)} \big[ r(x, y) \big],
\label{eq:rl}
\end{equation}
where $r(x, y) = v(x, y)$ in the basic setting.
Group relative policy optimization (GRPO)~\citep{DBLP:journals/corr/abs-2402-03300} samples $G$ responses $\{y^{(j)}\}_{j=1}^{G}$ per prompt and assigns every token of $y^{(j)}$ the group-normalized advantage
\begin{equation}
\hat{A}^{(j)} = \frac{r_j - \mathrm{mean}(\{r_k\}_{k=1}^{G})}{\mathrm{std}(\{r_k\}_{k=1}^{G})}, \qquad r_j \coloneqq r(x, y^{(j)}),
\label{eq:grpo-adv}
\end{equation}
which is optimized with a clipped surrogate objective.

\paragraph{Length-penalized reward.}
To control response length, the reward is commonly augmented with a length penalty~\citep{DBLP:journals/corr/abs-2607-24653, DBLP:journals/corr/abs-2609-23457}:
\begin{equation}
r(x, y) = v(x, y) - \lambda \, g\big( \Lth(y), B \big),
\label{eq:rl-reward}
\end{equation}
where $B$ is a target budget, $g(\cdot, \cdot) \ge 0$ is non-decreasing in $\Lth(y)$, and $\lambda \ge 0$ controls the strength of the penalty.
When training a single policy across multiple tiers $m \in \mathcal{M}$ jointly (\emph{Mixed RL}), each rollout under tier $m$ is scored with its corresponding budget $B_m$ and penalty weight $\lambda_m$.
In contrast to the per-token signal of OPD, \cref{eq:rl-reward} assigns a single scalar to the entire response.
\Cref{sec:phenomenology} instantiates both paradigms for the effort tiers in $\mathcal{M}$.

\section{Phenomenology: Compression, Controllability, and Generalization}
\label{sec:phenomenology}
\label{sec:discovery}
\label{sec:formulation}

A natural intuition for building an effort-controllable reasoning model is that multi-tier control is simply single-policy compression applied across multiple budgets: if a training objective can compress a policy to a target budget $B_m$, then conditioning a shared policy on $m \in \mathcal{M} = \{\low, \med, \high\}$ should suffice to separate effort tiers and transfer that control to new tasks. Does this intuition hold in practice? In this section, we test it by progressing from single-policy compression (\cref{sec:pheno-compress}) to joint multi-tier control within a shared policy (\cref{sec:pheno-joint,sec:pheno-opd}) and zero-shot transfer to unseen agentic formats (\cref{sec:pheno-transfer}), contrasting the \emph{sparse} trajectory-level reward of RL with the \emph{dense} token-level supervision of OPD at each step.

\begin{tcolorbox}[takeawaysbox]
\begin{itemize}[topsep=0pt,partopsep=0pt,leftmargin=12pt,itemsep=1.5pt]
\item \textbf{Sparse signals generalize compression across tasks but suffer from mode collapse.} Trajectory-level length penalties compress single-policy reasoning and generalize to unseen domains and tasks, but jointly training multiple tiers with sparse rewards collapses them to a single length or stalls the lower tiers (\cref{sec:pheno-compress,sec:pheno-joint}).
\item \textbf{Dense signals achieve multi-tier controllability on the formats covered in training.} State-wise targets in multi-teacher OPD separate effort tiers along the accuracy--compute frontier, and this separation is lost on multi-turn agentic tasks that are withheld from training (\cref{sec:pheno-opd,sec:pheno-transfer}).
\item \textbf{Compression and tier control transfer differently.} The shortening learned from sparse rewards carries over to unseen interaction formats, whereas the tier separation learned from dense targets follows the interaction formats seen in training.
\end{itemize}
\end{tcolorbox}

\subsection{Sparse Signals Compress a Single Policy}
\label{sec:pheno-compress}

\begin{figure}[!htb]
\centering
\includegraphics[width=\linewidth]{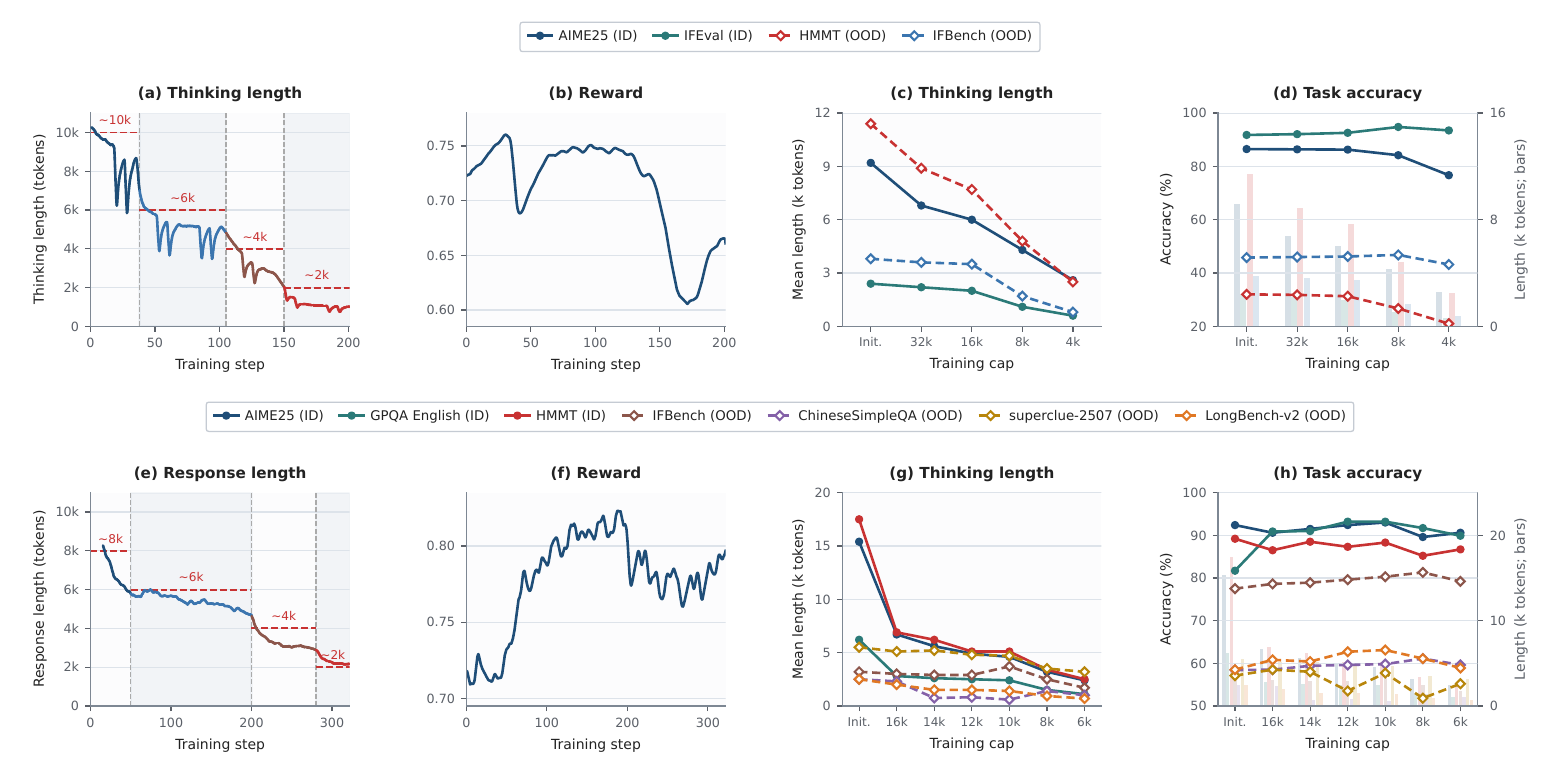}
\caption{\textbf{Single-policy compression with sparse length-penalized RL.} Top: Qwen3-8B~\citep{yang2025qwen3technicalreport}; bottom: GLM-4.5-Air~\citep{5team2025glm45agenticreasoningcoding}. (a,b,e,f)~Training length and reward. (c,g)~Evaluation thinking length across training caps. (d,h)~Task accuracy (lines) and thinking length (bars). Dataset legends identify ID and OOD benchmarks; Init. denotes the initial checkpoint.}
\label{fig:single_compress}
\end{figure}

We first examine single-policy compression: whether sparse trajectory-level length penalties induce learned shortening that persists under relaxed decoding caps and transfers to held-out domains. Directly imposing a tight decoding cap can truncate rollouts before they emit \eot, making shorter outputs insufficient evidence of learned stopping behavior. To reduce truncation during training, we use a \emph{staged overlong curriculum} that progressively tightens the active decoding cap, with the following buffered length penalty:
\begin{equation}
g\big(\Lth(y), B_s\big) = \min\!\left(1,\, \max\!\left(0,\, \frac{\Lth(y) - B_s}{L_{\mathrm{buf}}}\right)\right).
\label{eq:staged-cap-penalty}
\end{equation}
\paragraph{Setup.}
We train Qwen3-8B~\citep{yang2025qwen3technicalreport} and GLM-4.5-Air~\citep{5team2025glm45agenticreasoningcoding} with GSPO (\cref{eq:gspo}) and the length-penalized reward of \cref{eq:rl-reward} on single-turn mathematics, science, and instruction-following datasets. Across stages $s = 1, \dots, S$, we anneal the active decoding cap $L_{\max}^{(s)}$ from $16\text{k}$ to $6\text{k}$ on GLM-4.5-Air and from $32\text{k}$ to $4\text{k}$ on Qwen3-8B, with a linear penalty buffer $L_{\mathrm{buf}} = 4\text{k}$ ($B_s = L_{\max}^{(s)} - L_{\mathrm{buf}}$). All stage checkpoints are evaluated under a relaxed 32k decoding cap on both in-distribution (ID) benchmarks (AIME25, GPQA-Diamond, IFEval~\citep{dekoninck2026matharena,rein2024gpqa,DBLP:journals/corr/abs-2311-07911}) and held-out out-of-distribution (OOD) benchmarks excluded from training, tracking task accuracy, thinking length, and the cap-hit rate. Please refer to \cref{app:train-data,app:eval-protocol} for details.

\paragraph{Results.}
During training, mean thinking length decreases in stages, while mean reward fluctuates rather than improving monotonically (\cref{fig:single_compress}a,b). A separate checkpoint evaluation reveals two properties of the learned compression. First, \emph{shortening persists when the decoding constraint is relaxed}: checkpoints retain shortened thinking lengths under a 32k evaluation cap (\cref{fig:single_compress}c), supporting adaptation of the policy's stopping behavior. Second, \emph{shortening transfers to held-out domains while broadly preserving accuracy}: both in-distribution and held-out benchmarks show substantial length reductions with broadly stable accuracy over much of the compression schedule; degradation appears at the most aggressive settings (\cref{fig:single_compress}d). We next ask whether this ability to compress an individual policy extends to learning distinct effort tiers within a shared policy.

\subsection{Sparse Signals Fail at Multi-Tier Controllability}
\label{sec:pheno-joint}

\begin{figure}[!htb]
\centering
\includegraphics[width=\linewidth]{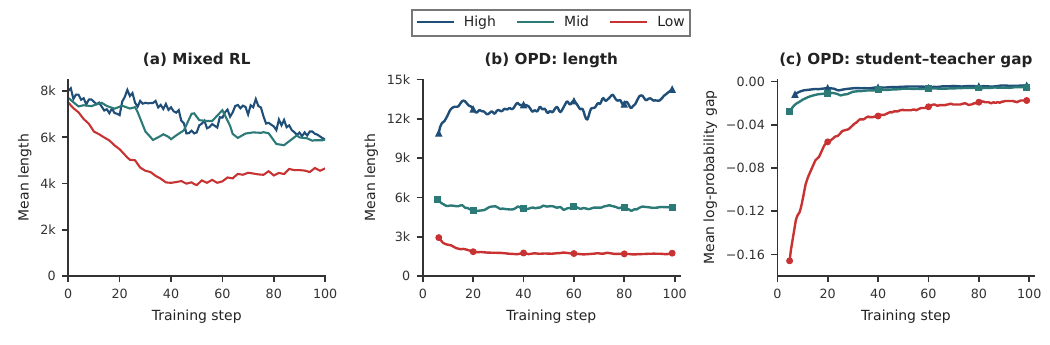}
\caption{\textbf{Failure modes of sparse Mixed RL vs.\ clean separation via dense OPD.} (a)~Unscaled sparse KL leads to cross-tier mode collapse. (b) and (c)~Dense OPD (\cref{sec:pheno-opd}) cleanly separates \high, \med, and \low length distributions with $<1\%$ truncation.}
\label{fig:joint}
\end{figure}

Given that sparse RL reliably compresses a single policy, a natural hypothesis is that conditioning a shared policy on $m \in \mathcal{M} = \{\low, \med, \high\}$ and applying tier-specific sparse rewards (Mixed RL) should suffice to separate effort tiers within a single model. To prevent lower-tier compression from overwriting \high, Mixed RL pairs sparse length penalties on \low and \med with a preserving sequence-level KL reward on \high, which exposes a tension in how competing trajectory-level scalars are balanced across shared parameters. We investigate whether scalar reward reweighting can resolve this tension and simultaneously preserve full reasoning on \high while compressing \low and \med.

\paragraph{Setup.}
We train GLM-4.5-Air (initialized from the 16k-cap compression checkpoint) across $m \in \{\low, \med, \high\}$ on the single-turn reasoning mixture under tier-specific decoding caps of 2k (\low), 4k (\med), and 16k (\high). Prompts on \low and \med receive the verifier-plus-length reward (\cref{eq:rl-reward}), while prompts on \high receive a sparse sequence-level KL reward toward $\pi_{\mathrm{init}}$. We compare two regimes: (i)~\emph{unscaled Mixed RL}, using the raw sequence-level KL reward on \high; and (ii)~\emph{advantage-rescaled Mixed RL}, which normalizes \high group advantages at each step to match the scale of the \low and \med length-penalized advantages. We track per-tier thinking length, cap-hit rate, and validation accuracy over 100 training steps. Please refer to \cref{app:train-data} for details.

\paragraph{Results.}
The two regimes show that preserving high-effort behavior and learning lower-effort stopping are separate requirements. First, \emph{joint shortening can erase conditional separation}: under unscaled Mixed RL, all three tiers converge to approximately 330 tokens by the end of training, accompanied by degraded reasoning accuracy (\cref{fig:joint}a). The policy becomes shorter, but the requested effort tier no longer distinguishes its mean thinking length. Second, \emph{preserving \high does not establish learned control of the lower tiers}: with rescaled high-tier advantages, \high remains at roughly 4--6k tokens, while \low stays near its 2k cap with more than $65\%$ truncation and \med fluctuates between 2.5k and 4.0k.
We analyze the objective-level constraints behind these outcomes in \cref{sec:mech-compete}. These results distinguish high-tier preservation from lower-tier compression: rescaling the high-tier advantages maintains long reasoning, but does not recover stable stopping behavior in the lower tiers.

\subsection{Dense Signals Achieve Multi-Tier Controllability}
\label{sec:pheno-opd}

\begin{figure}[!htb]
\centering
\includegraphics[width=\linewidth]{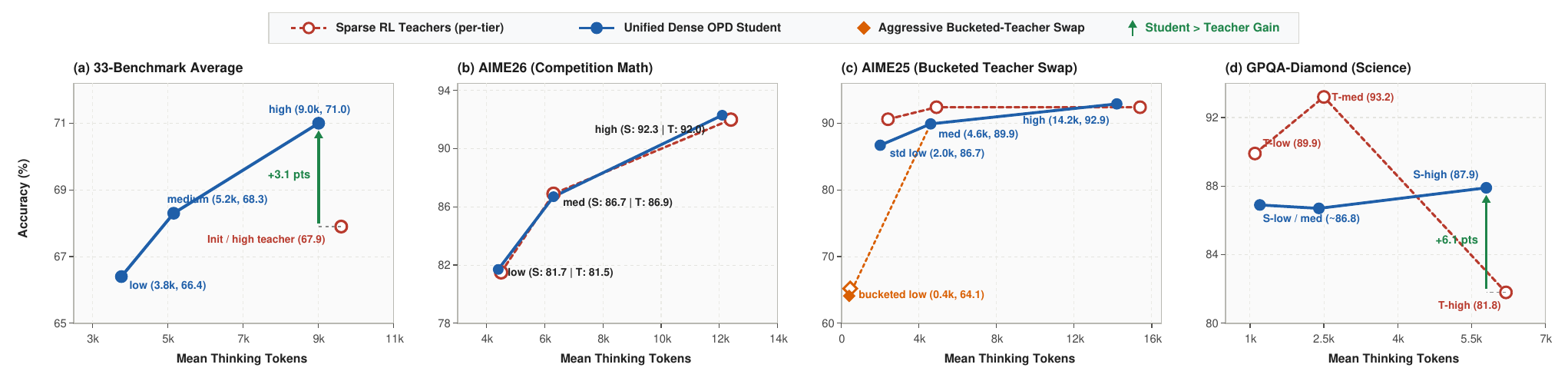}
\caption{\textbf{Multi-level effort control with OPD (GLM-4.5-Air~\citep{5team2025glm45agenticreasoningcoding}).} (a)~The unified student spans ordered accuracy--compute operating points. (b)~Student tiers track their teachers on AIME26~\citep{dekoninck2026matharena}. (c)~Replacing the \low teacher with a more compressed checkpoint shifts the corresponding student tier. (d)~The \high student improves GPQA-Diamond~\citep{rein2024gpqa} accuracy relative to its \high teacher.}
\label{fig:opd-frontier}
\end{figure}

Single-policy compression can produce teachers at different accuracy--compute operating points, but this does not by itself show that those behaviors can coexist in one effort-conditioned policy. We therefore next ask whether the resulting effort-specific behaviors can be consolidated into a shared policy while preserving their ordering and performance.

\paragraph{Setup.}
We obtain tier-specific teachers $\pi_T^m$ through single-policy compression (\cref{sec:pheno-compress}) and distill their next-token distributions with multi-teacher on-policy distillation (OPD; \cref{eq:opd})~\citep{DBLP:conf/iclr/AgarwalVZSGGB24,DBLP:journals/corr/abs-2606-30406}. We conduct the experiment on GLM-4.5-Air, initialize the student from the \high teacher, and route each effort request to its corresponding teacher. Training uses sampled-token reverse KL across mathematical, scientific, instruction-following, and agentic trajectories. We track response length and the student--teacher log-probability gap during training (\cref{fig:joint}b,c). The aggregate evaluation reports checkpoints at steps 15 and 20 across 33 benchmarks under a uniform 32k decoding cap, including mathematics, science, instruction-following, coding, and agentic tasks. For the AIME25 teacher-swap ablation, we replace the standard \low teacher ($2.4\text{k}$ thinking tokens) with a more compressed bucketed-penalty teacher ($0.4\text{k}$ tokens) while keeping the other teachers fixed. Please refer to \cref{app:train-data,app:benchmarks} for details.

\paragraph{Results.}
\emph{Dense consolidation yields distinct effort tiers within a shared policy.} The three response-length trajectories remain separated during training as the student--teacher log-probability gaps narrow (\cref{fig:joint}b,c). Under a common 32k evaluation cap, the student forms ordered low, medium, and high operating points, spanning 3.8k--9.0k thinking tokens and macro-average scores of $66.4$--$71.0$ (\cref{fig:opd-frontier}a).

\emph{Teacher replacement directly controls the corresponding tier.} Replacing the standard \low teacher with a $0.4\text{k}$ bucketed-penalty teacher moves the distilled \low tier to $0.4\text{k}$ tokens and $64.1\%$ accuracy, while the other tiers remain near their original operating points (\cref{fig:opd-frontier}c).

\emph{The high tier also improves in the aggregate evaluation.} The \high student reaches a macro-average score of $71.0$ versus $67.9$ for the reference policy, with $6.2\%$ fewer thinking tokens; on GPQA-Diamond, it improves from $81.8\%$ to $87.9\%$ (\cref{fig:opd-frontier}a,d).

In this setting, OPD produces ordered effort tiers while retaining a strong high-effort tier. Please refer to \cref{app:opd-separation} for the full results. The next section tests whether this ordering remains when multi-turn interaction data are withheld during training.

\subsection{Does Dense Controllability Generalize Out-of-Distribution?}
\label{sec:pheno-transfer}

\begin{figure}[!htb]
\centering
\includegraphics[width=\linewidth]{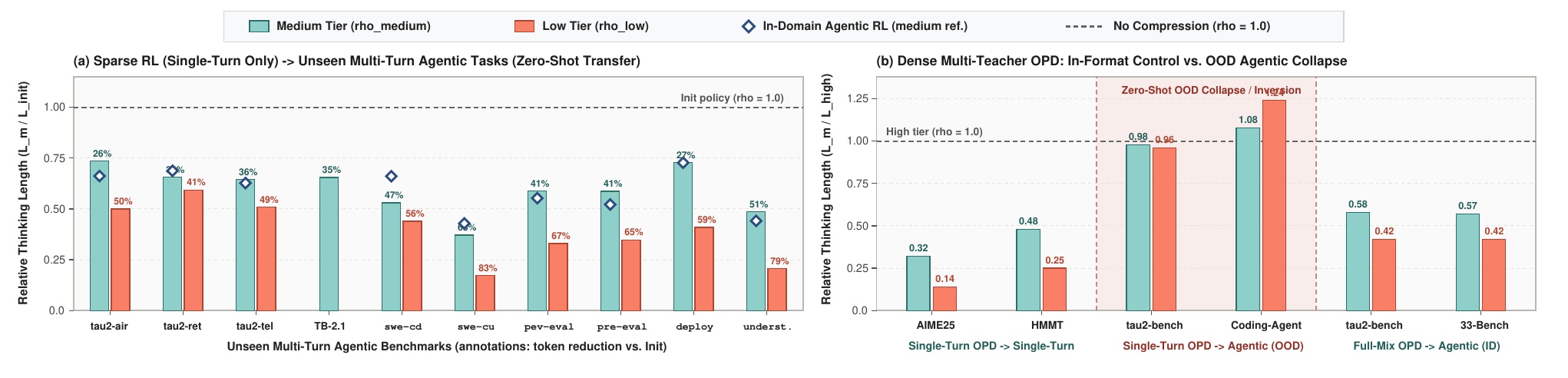}
\caption{\textbf{OOD format generalization: sparse RL vs.\ dense OPD.} (a)~Single-turn sparse RL compression reduces thinking length on unseen multi-turn agentic tasks while retaining accuracy. (b)~Single-turn dense OPD preserves tier separation on held-out single-turn math but collapses or inverts it on unseen agentic formats; including agentic data restores separation.}
\label{fig:pheno-transfer}
\end{figure}

The previous section established controllability when agentic trajectories are included in OPD training. Here we test whether that control transfers to a new interaction format: multi-turn tool use and coding-agent tasks that are withheld from training. We compare sparse RL compression and multi-teacher OPD under the same format shift.

\paragraph{Setup.}
Using GLM-4.5-Air, we train a \emph{reasoning-only} pipeline on single-turn mathematics, science, and instruction-following prompts, excluding tool calls and environment observations. We compare its sparse RL and multi-teacher OPD checkpoints with in-domain controls trained with agentic data. Evaluation covers held-out HMMT~\citep{dekoninck2026matharena} and ten multi-turn tool-use, terminal, and coding-agent tasks, using task accuracy and the relative length ratio $\rho_m = \Lth(m) / \Lth(\high)$. Please refer to \cref{app:train-data,app:benchmarks} for details.

\paragraph{Results.}
\emph{Sparse compression transfers across the format shift.} On the unseen agentic tasks, the reasoning-only \med and \low policies reduce thinking length by $43\%$ and $61\%$ on average without a systematic loss of task accuracy (\cref{fig:pheno-transfer}a).

\emph{Dense controllability remains format-dependent.} Single-turn OPD preserves effort ordering on held-out HMMT, with $\rho_{\med}=0.48$ and $\rho_{\low}=0.25$, but the same tiers collapse toward the high-tier length or invert on unseen agentic tasks ($\rho_m\approx0.96$--$0.98$ and $\rho_{\low}=1.24>\rho_{\med}=1.08$). Adding agentic trajectories restores separation, reaching $\rho_{\med}=0.58$ and $\rho_{\low}=0.42$ (\cref{fig:pheno-transfer}b).

This comparison separates two forms of transfer: sparse RL compression carries a stopping behavior across interaction formats, whereas dense OPD carries effort conditioning only when the relevant interaction states are represented during training. The objective-level constraints behind this difference are analyzed in \cref{sec:mechanism}. Please refer to \cref{app:rl-transfer} for the per-task results.

\begin{tcolorbox}[enhanced, colback=phenobg!60, colframe=phenoframe!35, boxrule=0.45pt, arc=1.8mm, borderline west={2.6pt}{0pt}{phenoframe}, fontupper=\small, left=3mm, right=3mm, top=1.6mm, bottom=1.6mm]
\textcolor{phenoframe}{\textbf{Core Insight (The Controllability--Generalization Duality).}}
\emph{Compression and tier control are governed by different signals and transfer differently:} sparse rewards learn a stopping behavior that carries over to unseen tasks and interaction formats but does not keep the tiers of a shared policy apart, whereas dense token-level targets separate the tiers, and the separation holds on the interaction formats covered in training.
\end{tcolorbox}

\section{Mechanism: What Governs Effort Controllability and Generalization?}
\label{sec:mechanism}
\label{sec:analysis}

\Cref{sec:phenomenology} left three observations unexplained: a shared policy trained with sparse length-penalized rewards does not keep its effort tiers apart (\cref{sec:pheno-joint}), dense token-level targets do (\cref{sec:pheno-opd}), and the control obtained from dense targets disappears on interaction formats that were absent from training (\cref{sec:pheno-transfer}). In this section, we trace all three to the \emph{granularity of the supervisory signal}, that is, to which decisions along a trajectory an objective actually constrains. We ask in turn why tiers collapse (\cref{sec:mech-compete}), why a single reward per response is not enough (\cref{sec:mech-dense}), and why control breaks on new tasks (\cref{sec:mech-transfer}). Please refer to \cref{app:proofs} for the proofs.

\begin{tcolorbox}[takeawaysbox]
\begin{itemize}[topsep=0pt,partopsep=0pt,leftmargin=12pt,itemsep=1.5pt]
\item \textbf{Tiers collapse because the shared objective is asymmetric.} Mixed RL penalizes length on \low and \med but contains no term that preserves \high, and the task reward of \high carries no gradient on groups with identical outcomes. Nothing in the objective opposes the shortening that reaches \high through the shared parameters (\cref{sec:mech-compete}).
\item \textbf{One reward per response under-determines where to shorten.} Rewards that score the reasoning process compress more than an answer-only reward at comparable accuracy, yet any aggregate is reliable only when the token-level signals it pools agree. Token-level targets remove the aggregation and constrain the stopping decision at every visited state (\cref{sec:mech-dense}).
\item \textbf{Dense control extends only to the states seen in training.} On agentic tasks, effort acts through the number of environment turns, a decision taken at states that single-turn training never visits. Token-level targets therefore leave it unconstrained (\cref{sec:mech-transfer}).
\end{itemize}
\end{tcolorbox}

\subsection{Why Do Tiers Collapse?}
\label{sec:mech-compete}

Single-policy compression succeeds with a sparse length penalty (\cref{sec:pheno-compress}), so it is not obvious why the same penalty should fail once three tiers share one policy (\cref{sec:pheno-joint}). We examine the Mixed RL objective directly and ask what, if anything, in that objective holds \high in place while \low and \med are shortened.

\paragraph{Setup.}
Mixed RL trains one policy $\pis$ on rollouts from all three tiers. A rollout at tier $m$ is scored with $r = v - \lambda_m\, g(\Lth(y), B_m)$ (\cref{eq:rl-reward}), where $\lambda_{\high}=0$ because \high is meant to keep its original reasoning length. The experiments of \cref{sec:pheno-joint} additionally give \high a sequence-level KL reward toward the initial policy; it is one scalar per rollout, and \cref{sec:mech-dense} examines why such a scalar does not preserve \high. Under GSPO (\cref{eq:gspo}), the clipped surrogate reduces locally, around $\theta_{\mathrm{old}}$, to
\begin{equation}
\mathcal{L}_{\mathrm{GSPO}}(\theta)
\approx -\E_{j}\!\left[\frac{\hat{A}^{(j)}}{|y^{(j)}|} \sum_{i=1}^{|y^{(j)}|} \log \pis\big(y_i^{(j)} \mid s_i, m\big)\right],
\label{eq:gspo-token-broadcast}
\end{equation}
so a single scalar, $\hat{A}^{(j)} / |y^{(j)}|$, weights every token of rollout $j$. Separating the task and length components of the reward, the population objective optimized by Mixed RL is
\begin{equation}
\mathcal{L}_{\mathrm{mixed}}(\theta)
= -\underbrace{\sum_{m \in \{\low,\med,\high\}} w_m\, \mathcal{J}^m_{\mathrm{task}}(\theta)}_{\text{task reward, all tiers}}
\;+\; \underbrace{\sum_{m \in \{\low,\med\}} w_m \lambda_m\, \mathcal{P}_m(\theta)}_{\text{length penalty, lower tiers only}},
\label{eq:mixed-asymmetric}
\end{equation}
where $\mathcal{J}^m_{\mathrm{task}}(\theta) = \E_{x,\, y \sim \pis^m}[v(x,y)]$, $\mathcal{P}_m(\theta) = \E_{x,\, y \sim \pis^m}[g(\Lth(y), B_m)]$, and $w_m \ge 0$ is the sampling weight of tier $m$.

\paragraph{Results.}
\emph{The objective contains no term that preserves \high.} Differentiating \cref{eq:mixed-asymmetric} gives
\begin{equation}
\nabla_\theta \mathcal{L}_{\mathrm{mixed}}
= -\sum_{m} w_m \nabla_\theta \mathcal{J}^m_{\mathrm{task}}
\;+\; \sum_{m \in \{\low,\med\}} w_m \lambda_m \nabla_\theta \mathcal{P}_m .
\label{eq:mixed-gradient}
\end{equation}
The second sum lowers the probability of continuing, and raises that of emitting \eot, on the prefixes visited by \low and \med. The three tiers differ only in the effort tag and otherwise share $\theta$, so any component of this update that does not depend on the tag acts on \high prefixes as well. \Cref{eq:mixed-gradient} has no length term for \high that could cancel it.

\emph{The task reward does not supply the missing restoring force.} In \cref{eq:mixed-asymmetric}, the only signal \high receives is its verifier outcome, and group normalization removes that signal whenever the rollouts of a group agree.

\begin{proposition}[Homogeneous groups carry no length gradient]
\label{prop:gspo-invariance}
Fix a prompt $x$ and an effort tier $m$, and let $r_j = v_j - \lambda g_j$ with $\lambda \ge 0$, where $g_j$ is the length penalty of rollout $j$. If all $G$ rollouts receive the same verifier outcome ($v_1 = \dots = v_G$), then
\[
\hat{A}^{(j)} = \frac{-\lambda\,(g_j - \bar{g})}{\lambda \sigma_g + \epsilon},
\]
where $\bar{g}$ and $\sigma_g$ are the group mean and standard deviation of the penalty. The advantage is zero when $\lambda = 0$ or $\sigma_g = 0$, and it converges to $-(g_j - \bar{g})/\sigma_g$, independent of $\lambda$, when $\lambda\sigma_g \gg \epsilon$.
\end{proposition}

For \high, $\lambda = 0$: on every prompt that its rollouts all solve, or all fail, the tier contributes no gradient, regardless of how far its reasoning length has drifted. When outcomes within a group differ, the advantage favors the correct rollouts irrespective of their length. In neither case does the update for \high depend on length, so shortening that does not change correctness goes unopposed.

\emph{Rescaling the penalty does not restore the separation.} \Cref{prop:gspo-invariance} also bounds what can be achieved by tuning $\lambda_m$ on the lower tiers. Once $\lambda_m \sigma_g \gg \epsilon$, the normalized advantage depends only on the within-group ranking of lengths, so a larger $\lambda_m$ does not strengthen the update. If instead all rollouts of a group saturate the penalty, as happens under the clipped penalty of \cref{eq:staged-cap-penalty} when every rollout overshoots its budget by more than $L_{\mathrm{buf}}$, then $\sigma_g = 0$ and the group carries no length signal at all. The penalty weight therefore sets neither how far apart the tiers are driven nor whether \high is retained.

Taken together, Mixed RL exerts shortening pressure through shared parameters and provides no opposing constraint on \high, which is the configuration illustrated in \cref{fig:joint}a. The \high ablations of Recipe I confirm the prediction empirically: without a preservation term, both the thinking length and the accuracy of \high deteriorate during training (\cref{sec:dense-definition}). The asymmetry is removed by adding the missing term,
\begin{equation}
\mathcal{L}_{\mathrm{anchor}}(\theta)
= \mathcal{L}_{\mathrm{mixed}}(\theta)
+ \beta\, \underbrace{\E_{x,\, y \sim \pis^{\high}}\!\left[\sum_{i=1}^{|y|} \KL\!\big(\pis^{\high}(\cdot \mid s_i) \,\big\|\, \pi^{\high}_{\mathrm{ref}}(\cdot \mid s_i)\big)\right]}_{\text{state-wise preservation of \high}},
\qquad \beta > 0,
\label{eq:anchor-completes-objective}
\end{equation}
which penalizes any change to the \high next-token distribution, including its stopping probability, at every state that \high visits. Whether the preservation term can itself be a sequence-level scalar is the subject of the next section.

\subsection{Why Is One Reward Not Enough?}
\label{sec:mech-dense}

\begin{figure}[!htb]
\centering
\includegraphics[width=\linewidth]{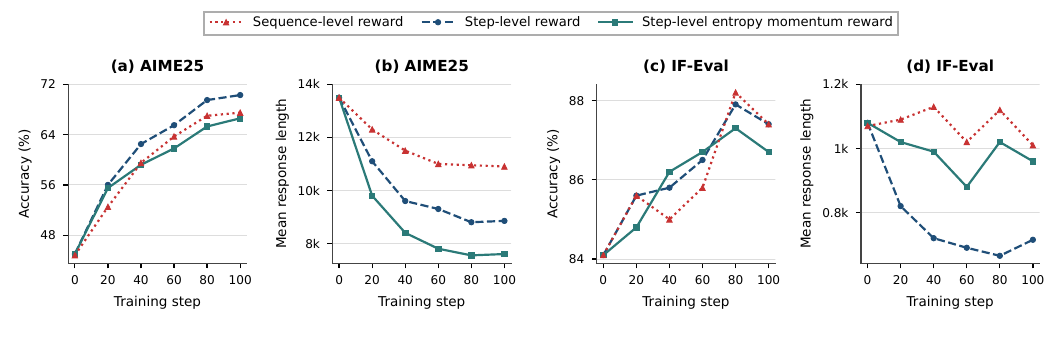}
\caption{\textbf{Answer-only versus process-aware rewards.} Accuracy and mean response length on AIME25~\citep{dekoninck2026matharena} (a,b) and IF-Eval~\citep{DBLP:journals/corr/abs-2311-07911} (c,d) over the first 100 training steps for a sequence-level (answer-only) reward and two step-level rewards. Both step-level rewards shorten AIME25 responses more than the sequence-level reward at comparable accuracy; on IF-Eval, only the step-level reward produces sustained shortening. Values are read from plotted training curves and are approximate.}
\label{fig:entropy-reward-comparison}
\end{figure}

\Cref{eq:gspo-token-broadcast} assigns one scalar to an entire response. The scalar records that a response was short or correct, but not which continuation decisions made it so. We ask whether this loss of information matters in practice, first by enriching the reward with information about the reasoning process, and then by examining how far the aggregation of a token-level signal can be coarsened before it stops being informative.

\begin{figure}[!htb]
\centering
\includegraphics[width=\linewidth]{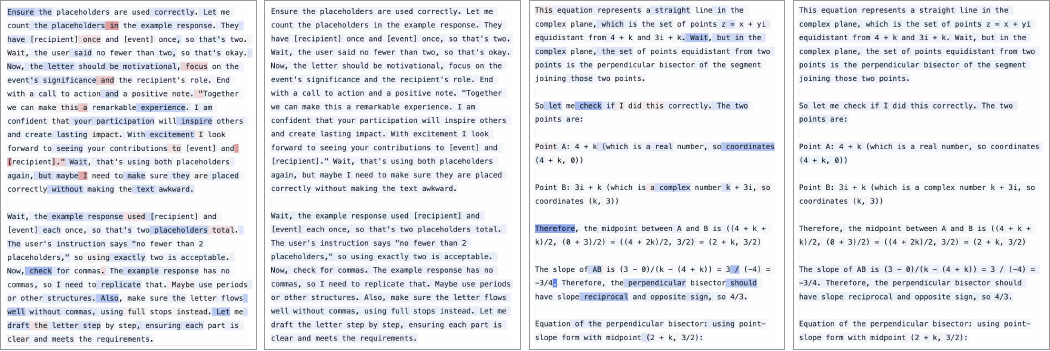}
\caption{\textbf{Token-level and step-level student--teacher gaps on the same responses.} From left to right: token-level and step-level gaps on an instruction-following response, and token-level and step-level gaps on a mathematics response. Shading marks the gap at each token. Within one instruction-following step, token-level gaps point in different directions and step-level aggregation replaces them with a single value; within a mathematics step, the gaps are largely aligned.}
\label{fig:mech-gap-variance}
\end{figure}

\paragraph{Setup.}
We conduct two analyses on Qwen3-8B~\citep{yang2025qwen3technicalreport}, both trained on a mixture of AIME25 and IFEval prompts~\citep{dekoninck2026matharena,DBLP:journals/corr/abs-2311-07911}. \emph{(1) Reward granularity.} We compare three GSPO rewards (\cref{fig:entropy-reward-comparison}). The \emph{sequence-level reward} is the verifier outcome of the final answer. The two \emph{step-level rewards} additionally score the reasoning process: the thinking trace is split into steps at blank lines, the mean token entropy $H_t$ of each step is tracked through the momentum $R_t = \gamma R_{t-1} + \alpha \max(0, H_{t-1} - H_t) - \beta \max(0, H_t - H_{t-1})$, and each step of a correct response receives a bonus that increases with its standardized momentum. The \emph{step-level reward} standardizes within a trajectory, and the \emph{step-level entropy momentum reward} standardizes across the correct responses to the same prompt. Step scores are averaged into the reward of the response, so the update still uses one scalar per rollout, but that scalar now distinguishes among correct rollouts. \emph{(2) Aggregation granularity.} We add the OPD advantage of a length-compressed teacher to GSPO and compare the token-level form $A_i = -\ell_i$ (\cref{eq:opd-sample}) with a step-level form in which every token of a step receives the mean of $A_i$ over that step. We inspect the sampled-token gaps $\ell_i$ on individual responses and measure their variance within a step (\cref{fig:mech-gap-variance}).

\paragraph{Results.}
\emph{Process information determines how much a sparse reward compresses.} With the sequence-level reward, mean response length on AIME25 falls from about $13.5$k to $10.9$k tokens over 100 steps, whereas the step-level reward and its entropy momentum variant reach about $8.9$k and $7.6$k tokens (\cref{fig:entropy-reward-comparison}b). Accuracy rises from about $45\%$ to $67.5\%$, $70.3\%$, and $66.6\%$, respectively (\cref{fig:entropy-reward-comparison}a). On IF-Eval, the sequence-level reward shows no consistent shortening, the step-level reward shortens responses from about $1.08$k to $0.72$k tokens, and the entropy momentum variant reaches about $0.96$k tokens, with accuracy within one point across the three (\cref{fig:entropy-reward-comparison}c,d). This pattern follows from \cref{prop:gspo-invariance}: under an answer-only reward, a group of uniformly correct rollouts has zero advantage, so nothing favors its shorter members, whereas a process-aware reward breaks the tie. The amount of compression nonetheless depends on the task and on how the step scores are normalized.

\emph{An aggregate is informative only where the token-level signals it pools agree.} On mathematics, the step-level form of the OPD advantage shortens reasoning. On instruction following, the token-level form keeps the length under control, while under the step-level form the length rises again during training. \Cref{fig:mech-gap-variance} shows the reason. A single instruction-following step interleaves constraint planning, counting, content generation, and re-checking, and the token-level gaps within it take opposite signs. Averaging replaces them with one value that reflects none of the individual decisions. A mathematics step is typically one coherent derivation whose gaps share a direction, so its mean remains representative. The within-step variance of the token-level gap is $0.537$ in the reasoning trace and $1.388$ in the answer for instruction following, against $0.183$ and $0.208$ for mathematics. A sequence-level scalar is the limiting case of this aggregation, pooling every decision of the response into one number.

\emph{Token-level targets constrain the stopping decision itself.} Whether to continue reasoning or to stop is decided at individual states, and the exact OPD objective places a constraint on each of them.

\begin{proposition}[Dense distillation supervises stopping at every state]
\label{prop:opd-stopping}
Let $e = \eot$ and let $\dB(a \,\|\, b) = a \log \frac{a}{b} + (1{-}a) \log \frac{1{-}a}{1{-}b}$ denote the Bernoulli KL divergence. For the student and teacher next-token distributions $p_i$ and $q_i$ at any state $s_i$,
\begin{equation}
\KL(p_i \,\|\, q_i) \;\ge\; \dB\big(p_i(e) \,\big\|\, q_i(e)\big),
\label{eq:bernoulli-stopping-bound}
\end{equation}
and hence
\begin{equation}
\mathcal{L}^m_{\mathrm{OPD}}(\theta) \;\ge\; \E_{x,\, y \sim \pis^m}\!\left[\sum_{i \le \min(\Lth(y)+1,\, |y|)} \dB\!\big(\pis^m(e \mid s_i) \,\big\|\, \pi_T^m(e \mid s_i)\big)\right].
\label{eq:opd-stopping-bound}
\end{equation}
\end{proposition}

Driving the OPD loss toward zero therefore forces the student's probability of ending its reasoning toward that of the tier-specific teacher at every state it visits. Since the teachers $\pi_T^{\low}$, $\pi_T^{\med}$, and $\pi_T^{\high}$ assign different stopping probabilities to the same prefix, the tiers are held apart state by state instead of through a trajectory-level length statistic. The OPD dynamics of \cref{sec:pheno-opd} are consistent with this account: the three response lengths remain separated while the student--teacher log-probability gaps narrow (\cref{fig:joint}b,c). For the anchor in \cref{eq:anchor-completes-objective}, the same reasoning implies that the preservation term must be applied per token. A trajectory-level KL reward is again a single scalar per rollout, and in Recipe I it fails to prevent the \high tier from shortening (\cref{sec:dense-definition}).

\subsection{Why Does Control Break on New Tasks?}
\label{sec:mech-transfer}

\Cref{sec:mech-dense} showed that token-level targets constrain the stopping decision at every state the student visits. Tiers distilled from single-turn data nevertheless lose their separation on agentic tasks (\cref{sec:pheno-transfer}). We first locate where the control is lost, and then examine what an effort request has to change on those tasks.

\paragraph{Setup.}
We distill three effort tiers with multi-teacher OPD using single-turn reasoning data only, and evaluate the students together with the high-effort policy that initializes distillation on 16 benchmarks: reasoning, which the distillation data cover, and general, code, and agent benchmarks, which they do not (\cref{tab:opd-transfer-scope}). To see how effort acts on multi-turn tasks, we decompose agentic trajectories into the number of environment turns and the reasoning length per turn, using roughly $3.2$k trajectories per effort tier from a large MoE reasoning model (\cref{tab:agent-turns}).

% Cells marked \fillme are thinking lengths still to be filled in (replace \fillme with the value, e.g. 12.3k).
\providecommand{\fillme}{\textcolor{accent}{--}}
\begin{table}[!htb]
\centering
\footnotesize
\setlength{\tabcolsep}{3.2pt}
\renewcommand{\arraystretch}{1.05}
\begin{tabular}{@{}c l r rrr r r rrr r@{}}
\toprule
 & & \multicolumn{5}{c}{\textbf{Task score}} & \multicolumn{5}{c}{\textbf{Thinking tokens}} \\
\cmidrule(lr){3-7}\cmidrule(l){8-12}
 & \textbf{Benchmark} & \textbf{Baseline} & \textbf{\high} & \textbf{\med} & \textbf{\low} & \textbf{\high$-$\low} & \textbf{Baseline} & \textbf{\high} & \textbf{\med} & \textbf{\low} & \textbf{\high$-$\low} \\
\midrule
\multirow{7}{*}{\rotatebox[origin=c]{90}{\textbf{Reasoning}}} & AIME26 & 93.02 & 92.71 & 90.83 & 86.77 & \cellcolor{best!14}$+5.94$ & 24.9k & 25.7k & 13.4k & 7.8k & +17.9k \\
 & HMMT-2602 & 85.89 & 85.70 & 81.06 & 73.77 & \cellcolor{best!30}$+11.93$ & 28.3k & 29.2k & 17.1k & 9.9k & +19.3k \\
 & GPQA-Diamond & 91.41 & 89.90 & 90.03 & 84.97 & \cellcolor{best!14}$+4.93$ & 11.0k & 12.5k & 5.2k & 2.7k & +9.8k \\
 & HLE & 26.41 & 25.72 & 21.46 & 17.89 & \cellcolor{best!14}$+7.83$ & 24.2k & 26.8k & 12.3k & 6.7k & +20.1k \\
 & ARC-AGI-1 & 77.57 & 79.47 & 70.64 & 65.39 & \cellcolor{best!30}$+14.08$ & 23.4k & 22.7k & 15.9k & 11.1k & +11.6k \\
 & ARC-AGI-2 & 16.17 & 15.57 & 11.38 & 7.78 & \cellcolor{best!14}$+7.79$ & 43.5k & 43.1k & 38.1k & 3.5k & +39.6k \\
 & \textbf{Average} & \textbf{65.08} & \textbf{64.84} & \textbf{60.90} & \textbf{56.09} & \cellcolor{best!30}$\mathbf{+8.75}$ & \textbf{25.9k} & \textbf{26.7k} & \textbf{17.0k} & \textbf{7.0k} & \textbf{+19.7k} \\
\midrule
\multirow{5}{*}{\rotatebox[origin=c]{90}{\textbf{General}}} & AA-LCR & 71.33 & 69.67 & 68.00 & 68.67 & \cellcolor{accent!12}$+1.00$ & 4.3k & 4.4k & 1.8k & 1.1k & +3.3k \\
 & LongBench-v2& 65.00 & 66.54 & 63.46 & 65.77 & \cellcolor{accent!12}$+0.77$ & 5.3k & 5.3k & 2.5k & 1.2k & +4.1k \\
 & IFEval & 87.43 & 87.43 & 87.25 & 88.72 & \cellcolor{accent!28}$-1.29$ & 2.5k & 2.9k & 1.6k & 1.2k & +1.7k \\
 & IFBench & 60.88 & 61.22 & 57.48 & 56.12 & \cellcolor{best!14}$+5.10$ & 7.3k & 7.6k & 4.9k & 3.5k & +4.1k \\
 & \textbf{Average} & \textbf{71.16} & \textbf{71.22} & \textbf{69.05} & \textbf{69.82} & \cellcolor{accent!12}$\mathbf{+1.40}$ & \textbf{4.8k} & \textbf{5.0k} & \textbf{2.7k} & \textbf{1.8k} & \textbf{+3.3k} \\
\midrule
\multirow{3}{*}{\rotatebox[origin=c]{90}{\textbf{Code}}} & SWE-Compass & 88.00 &81.00 & 81.00 & 78.50 & \cellcolor{accent!12}$+2.50$ & 79.6k & 80.8k & 71.6k & 76.3k & +4.5k \\
 & Terminal-Bench 2.1 & 62.70 & 57.75 & 60.00 & 56.85 & \cellcolor{accent!12}$+0.90$ & 40.8k & 36.5k & 34.2k & 34.8k & +1.7k \\
 & \textbf{Average} & \textbf{75.35} & \textbf{69.38} & \textbf{70.50} & \textbf{67.67} & \cellcolor{accent!12}$\mathbf{+1.70}$ & \textbf{60.2k} & \textbf{58.6k} & \textbf{52.9k} & \textbf{55.5k} & \textbf{+3.1k} \\
\midrule
\multirow{5}{*}{\rotatebox[origin=c]{90}{\textbf{Agent}}} & Pev-Bench & 87.70 & 86.05 & 85.61 & 87.70 & \cellcolor{accent!28}$-1.65$ & 22.5k & 23.3k & 19.6k & 17.2k & +6.1k \\
 & $\tau^2$-retail & 90.70 & 92.11 & 92.11 & 92.63 & \cellcolor{accent!28}$-0.52$ & 22.1k & 22.2k & 22.6k & 20.5k & +1.7k \\
 & $\tau^2$-airline & 86.00 & 85.60 & 85.60 & 84.80 & \cellcolor{accent!12}$+0.80$ & 16.8k & 17.2k & 16.9k & 16.6k & +0.6k \\
 & VitaBench & 62.06 & 61.00 & 62.91 & 61.75 & \cellcolor{accent!28}$-0.75$ & 14.1k & 13.2k & 13.0k & 12.4k & +0.8k \\
 & \textbf{Average} & \textbf{81.62} & \textbf{81.19} & \textbf{81.56} & \textbf{81.72} & \cellcolor{accent!28}$\mathbf{-0.53}$ & \textbf{18.9k} & \textbf{19.0k} & \textbf{18.0k} & \textbf{16.7k} & \textbf{+2.3k} \\
\bottomrule
\end{tabular}
\caption{\textbf{Task scores and thinking lengths of OPD students distilled on single-turn reasoning data only.} The baseline is the high-effort policy that initializes distillation, and the \high$-$\low columns give the gap between the \high and \low students. For the score gap, blue cells mark at least three points, darker from eight points; red cells mark less than three points, darker where \low scores at least as high as \high. A red \high score marks a loss of at least three points relative to the baseline.}
\label{tab:opd-transfer-scope}
\end{table}

\paragraph{Results.}
\emph{Control holds on the single-turn format and is lost on multi-turn tasks.} On the six reasoning benchmarks, the \high student retains the baseline score ($64.8$ vs.\ $65.1$ on average) and the tiers are ordered in both length and score: \med and \low use $64\%$ and $26\%$ of the \high thinking tokens and score $3.9$ and $8.8$ points lower, with a gap between \high and \low of at least $4.9$ points on every benchmark. On the code and agent benchmarks, the effort request has little effect. \low still uses $95\%$ and $88\%$ of the \high thinking tokens, and the score gap between \high and \low is below three points on all six benchmarks. On code, the \high student also scores $6.0$ points below the baseline on average. The general benchmarks separate the two candidate causes. Their tasks are absent from the distillation data, as code and agent tasks are, yet they are single-turn, as the reasoning data are, and on them thinking length remains ordered: \med and \low use $54\%$ and $36\%$ of the \high length. Control therefore follows the interaction format, not the task domain.

\begin{table}[!htb]
\centering\small
\caption{Agentic trajectories of a multi-level DeepSeek-V4-Flash model. Effort
changes the number of turns, not the reasoning per turn.}
\label{tab:agent-turns}
\begin{tabular}{@{}lccc@{}}
\toprule
 & \low & \med & \high \\
\midrule
Trajectories & 3{,}212 & 3{,}148 & 3{,}256 \\
Turns per trajectory (mean / median) & 78.2 / 70 & 85.8 / 76 & 91.6 / 84 \\
Reasoning tokens per turn (mean / median) & 59.1 / 19 & 62.1 / 20 & 60.9 / 19 \\
Reasoning tokens per trajectory (mean) & 4{,}618 & 5{,}324 & 5{,}575 \\
First-turn reasoning tokens (mean) & 179.9 & 189.6 & 197.8 \\
\bottomrule
\end{tabular}
\end{table}

\emph{On multi-turn tasks, effort is exercised through the number of turns.} From \high to \low, reasoning tokens per trajectory fall by $17.2\%$ (from $5{,}575$ to $4{,}618$). The mean number of turns falls by $14.6\%$ (from $91.6$ to $78.2$), whereas reasoning per turn changes by only $3.0\%$ (from $60.9$ to $59.1$ tokens, with a median of $19$ at both tiers). Nearly all of the difference between tiers lies in whether the model takes another action after an environment observation, and very little in how long it reasons before each action.

\emph{Single-turn training never visits the states at which this decision is made.} A single-turn trace consists of a prompt followed by one reasoning segment and an answer. It contains no tool observation, and hence no state at which the choice between acting again and submitting arises. The expectation in \cref{eq:opd-stopping-bound} runs over the states that the student visits during distillation, so the tier-specific teachers constrain when to stop reasoning within a turn and place no constraint on whether to take another turn. The first decision transfers to any single-turn task, which is why length stays ordered on the general benchmarks. The second is the one that governs effort on code and agent tasks, and it is left to the initialization, which is shared by the three tiers. Including agentic trajectories in distillation brings these states into the expectation and restores the separation (\cref{fig:pheno-transfer}b).

\emph{Sparse compression is not tied to states in the same way.} Single-turn sparse RL shortens reasoning on unseen agentic tasks by $43\%$ and $61\%$ on average without a systematic loss of accuracy. A trajectory-level penalty does not prescribe a target at particular states; it shifts the policy toward earlier stopping in general, and this shift carries over to new formats. The same lack of state-specific targets is what prevents it from separating tiers (\cref{sec:mech-compete}). Please refer to \cref{app:rl-transfer} for the per-task results.

\begin{tcolorbox}[enhanced, colback=phenobg!60, colframe=phenoframe!35, boxrule=0.45pt, arc=1.8mm, borderline west={2.6pt}{0pt}{phenoframe}, fontupper=\small, left=3mm, right=3mm, top=1.6mm, bottom=1.6mm]
\textcolor{phenoframe}{\textbf{Mechanistic Synthesis (From Diagnosis to Training Design).}}
\begin{itemize}[leftmargin=11pt, topsep=2pt, partopsep=0pt, itemsep=1.5pt]
\item \textbf{Preserve \high with a token-level anchor.} Tier collapse stems from a missing preservation term, not from the length penalty on the lower tiers. A state-wise KL anchor on \high supplies that term inside a single RL stage, with no need for a ladder of separately trained teachers (\cref{sec:mech-compete,sec:mech-dense}).
\item \textbf{Keep sparse rewards for compression.} Length-penalized rewards shorten \low and \med and carry that shortening across formats; process-aware rewards strengthen the effect where answer-only rewards cannot rank correct rollouts (\cref{sec:mech-dense,sec:mech-transfer}).
\item \textbf{Make the effort request meaningful on unseen states.} Dense targets control effort only on visited states, so control on a new format requires either coverage of that format or an effort specification whose meaning does not have to be learned from those states (\cref{sec:mech-transfer}).
\end{itemize}
\end{tcolorbox}

\section{Practical Recipes}
\label{sec:dense-rl}
\label{sec:recipe}
\label{sec:generalization}
\label{sec:transfer-results}
\label{sec:recipe-scope}

The analysis in \cref{sec:mechanism} explains why multi-stage teacher training followed by distillation succeeds in \cref{sec:pheno-opd}, and it also shows that the teacher cascade is not required for multi-tier effort control. \Cref{sec:mech-compete,sec:mech-dense} trace tier collapse in Mixed RL to a missing state-wise preservation term for \high, which implies that a token-level KL anchor on \high should keep the tiers apart within a single RL stage. \Cref{sec:mech-transfer} shows that dense effort control extends to a new interaction format only if training covers that format or the effort specification carries a meaning that does not have to be learned from its states. In this section, we turn these two principles into single-stage RL recipes: \textbf{Recipe I: Single-Stage Dense-Anchor RL} (\cref{sec:dense-definition}), which replaces multi-stage OPD with a per-token KL anchor on \high and uses only discrete effort tags $m \in \{\low, \med, \high\}$; and \textbf{Recipe II: Prompt-Guided Dense-Anchor RL} (\cref{sec:prompt-opd}), which adds tier-specific natural-language effort instructions, so that the tiers separate within 20 training steps and the separation carries over to multi-turn code and agent benchmarks without agentic training data.

\begin{tcolorbox}[takeawaysbox]
\begin{itemize}[topsep=0pt,partopsep=0pt,leftmargin=12pt,itemsep=1.5pt]
\item \textbf{A single high-tier token-level KL anchor enables single-stage multi-tier RL.} Anchoring only \high at the token level prevents the lower-tier length penalties from shortening \high through the shared parameters, eliminating the need for multi-stage teacher training (\cref{sec:dense-definition}).
\item \textbf{Natural-language effort prompts unlock rapid convergence and zero-shot format transfer.} Replacing bare discrete tags with natural-language effort instructions leverages the base model's instruction-following prior, enabling fast single-turn RL to generalize zero-shot to unseen multi-turn agents (\cref{sec:prompt-opd}).
\end{itemize}
\end{tcolorbox}

\subsection{Recipe I: Dense-Anchor RL}
\label{sec:dense-definition}
\label{sec:dense-dynamics}
\label{sec:dense-id}
\label{sec:dense-single}

Mixed RL (\cref{sec:pheno-joint}) applies tier-specific length penalties to one shared policy. When rollouts for the \high tier receive the same correctness outcome, \cref{prop:gspo-invariance} implies that their relative update is determined primarily by length differences. At the same time, the length penalties for \low and \med favor shorter traces. This creates a conflict: the lower tiers are encouraged to stop earlier, while the sparse objective provides no dense signal that preserves the behavior of \high. Recipe I tests whether a token-level preservation term can resolve this conflict within a single RL stage.

\begin{figure}[!htb]
\centering
\includegraphics[width=0.92\linewidth]{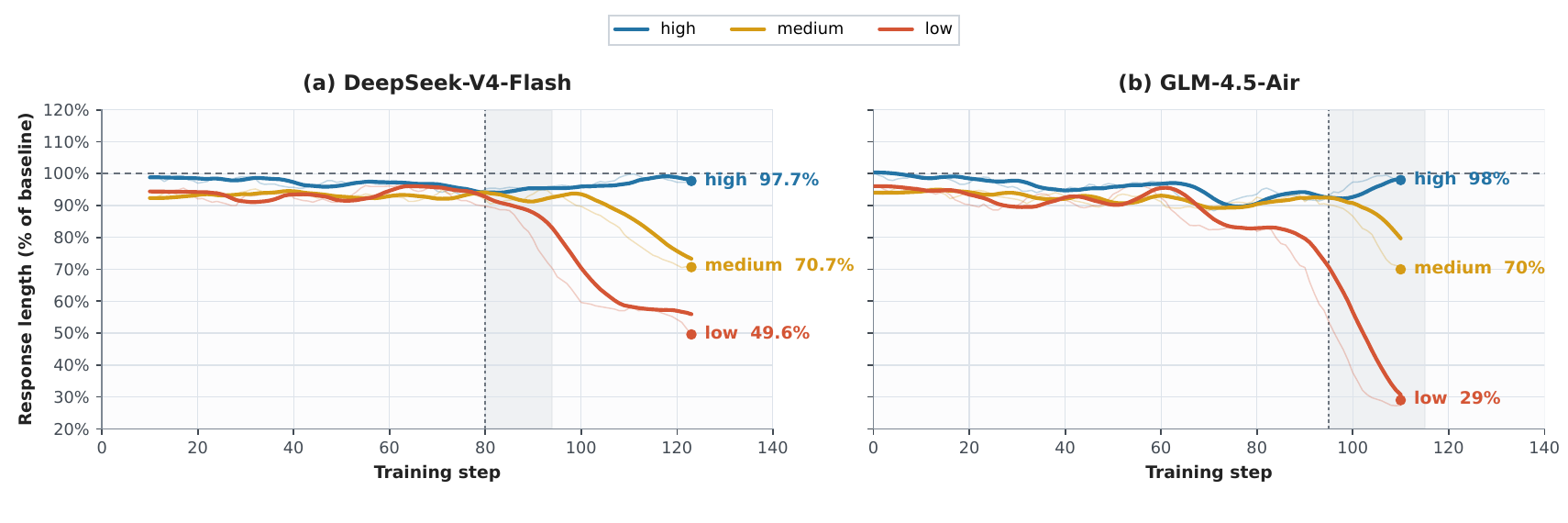}
\caption{\textbf{Emergence of effort control during Recipe I training.} Curves show 10-step weighted averages of response length relative to the initialization checkpoint. The shaded regions mark the reported training transition. The final labels give the reported length ratios.}
\label{fig:recipe1-emergence}
\end{figure}

\paragraph{Setup.}
The policy receives a discrete effort tag $m \in \{\low, \med, \high\}$. The \low and \med tiers are optimized with length-penalized RL, while the \high tier is anchored to the frozen initialization policy $\pi_{\mathrm{ref}}$ through a token-level KL penalty:
\begin{equation}
\mathcal{L}_{\mathrm{DA\text{-}RL}}(\theta) = -\textstyle\sum_{m \in \{\low, \med\}} \mathcal{J}^m_{\mathrm{RL}}(\theta) \;+\; \beta\,\E_{x,\, y \sim \pis^{\high}} \!\left[\textstyle\sum_{i=1}^{|y|} \hat{k}\big(\pis^{\high}(y_i \mid s_i),\, \pi_{\mathrm{ref}}^{\high}(y_i \mid s_i)\big)\right],
\label{eq:dense-rl}
\end{equation}
where $\hat{k}$ is a per-token KL estimator and $\beta=0.5$. The comparison includes two alternatives for the \high tier: removing the preservation term and replacing it with a trajectory-level KL reward. All training uses single-turn data.

\paragraph{Results.}
\textbf{Token-level anchoring preserves the high-effort tier and eventually separates the lower tiers.}
Without a preservation term, both the thinking length and task performance of \high deteriorate during training. A trajectory-level KL reward also fails to prevent this collapse: because its penalty accumulates over the generated trace, shorter responses incur a smaller penalty, and the thinking length decreases from approximately $20$k to $1$k tokens within 30 steps. With the token-level anchor in \cref{eq:dense-rl}, all three tiers remain within roughly $10\%$ of the reference length for about 80 steps before separating (\cref{fig:recipe1-emergence}). After 123 steps, the \high, \med, and \low tiers reach $97.7\%$, $70.7\%$, and $49.6\%$ of the reference length, respectively; a second run reaches $98\%$, $70\%$, and $29\%$.

\textbf{Separation is slow and transfer is limited.}
The extended period before the tiers separate increases the training cost. The effort tags are distinguishable on the single-turn tasks used for training, but no clear separation is observed on agent tasks. Thus, the discrete tags can support effort control, but their meaning must be learned through RL rather than being specified by the input.

\begin{tcolorbox}[takeawaysbox]
Recipe I provides evidence for the mechanism in \cref{sec:mechanism}: \textbf{\textcolor{Purple}{a dense token-level signal can produce ordered effort tiers within a single RL stage}}. The result has two limitations. Separation takes many training steps, and the learned control transfers weakly beyond the single-turn training format. These limitations motivate Recipe II, which supplies natural-language effort instructions so that the intended reasoning depth is specified in the input rather than inferred solely from discrete tags.
\end{tcolorbox}

\subsection{Recipe II: Prompt-Guided Dense-Anchor RL}
\label{sec:prompt-opd}

The limitations in Recipe I (\cref{sec:dense-definition}) raise a question: how can we accelerate effort separation and improve generalization beyond the training format? Instruction following suggests a possible approach: describing the intended reasoning depth in natural language rather than using discrete flags alone. In this section, we explore this approach by pairing each effort tier $m$ with an instruction $c_m$ during Dense-Anchor RL. We test whether this guidance accelerates effort separation and improves transfer to held-out task formats, and compare the resulting policy with multi-teacher OPD.

\paragraph{Algorithm.}
For each effort tier $m \in \{\low,\,\med,\,\high\}$, we prepend a
tier-specific instruction $c_m$ to the prompt and optimize the resulting
instruction-conditioned policy. The training objective is
\begin{equation}
\begin{aligned}
\mathcal{L}_{\mathrm{PG\text{-}DA\text{-}RL}}(\theta)
={}&-\sum_{m \in \{\low,\,\med\}}
    \mathcal{J}^m_{\mathrm{GSPO}}(\theta; c_m) \\
&+\beta\,\E_{x,\, y \sim \pis(\cdot \mid x, c_{\high}, \high)}
\!\left[\sum_{i=1}^{|y|}
\hat{k}\!\left(\pis(y_i \mid s_i, c_{\high}, \high),\,
\pi_{\mathrm{ref}}(y_i \mid s_i, c_{\high}, \high)\right)\right].
\end{aligned}
\label{eq:prompt-opd}
\end{equation}
The GSPO term updates the \low and \med policies under their respective instructions. The second term is a dense token-level KL anchor for \high: its trajectories are sampled under $c_{\high}$, and deviations from the frozen reference policy are penalized at every generation step. This preserves the high-effort behavior while allowing the lower-effort policies to adapt their response lengths.

\paragraph{Prompt Template.}
The effort instructions are listed below. They specify the intended reasoning depth while leaving the task prompt and answer format unchanged. Each training example is paired with the instruction corresponding to its target effort tier.

\begin{tcolorbox}[
    enhanced, colback=black!2, colframe=black!20,
    boxrule=0.4pt, arc=0.8mm,
    colbacktitle=opdhigh!8, coltitle=opdhigh!85!black,
    title={Effort Instruction Templates},
    fonttitle=\small\bfseries, fontupper=\small,
    titlerule=0.4pt,
    left=3mm, right=3mm, top=2mm, bottom=2mm,
    before skip=8pt, after skip=8pt]
\noindent\textcolor{opdhigh}{\textbf{High-effort prompt template}}\hfill $c_{\high}$\par\smallskip
\noindent You are in high-intensity deep thinking mode. Before answering, conduct comprehensive, in-depth, multi-angle reasoning and deduction. Proactively decompose complex problems, thoroughly consider edge cases, implicit conditions, and counterexamples, and perform rigorous self-verification and reflection before concluding.
\par\medskip
{\color{black!15}\hrule height 0.4pt}
\medskip
\noindent\textcolor{opdhigh}{\textbf{Medium-effort prompt template}}\hfill $c_{\med}$\par\smallskip
\noindent You are in standard thinking mode. Before answering, develop clear and structured reasoning around the core problem. Maintain moderate depth, focus directly on key logical steps, and avoid excessive digression or repetitive verification of simple details.
\par\medskip
{\color{black!15}\hrule height 0.4pt}
\medskip
\noindent\textcolor{opdhigh}{\textbf{Low-effort prompt template}}\hfill $c_{\low}$\par\smallskip
\noindent You are in minimal fast thinking mode. Think concisely only about the most essential steps, quickly lock onto key information, and cut straight to the conclusion. Omit tedious intermediate derivations, background setup, and repeated checks.
\end{tcolorbox}

\paragraph{Setup.}
DeepSeek-V4-Flash~\citep{DBLP:journals/corr/abs-2606-19348} and GLM-4.5-Air~\citep{5team2025glm45agenticreasoningcoding} are trained using GSPO for 20 and 25 steps, respectively, with a batch size of 256, a learning rate of $10^{-6}$, and a dense-anchor coefficient of $\beta=0.5$. The training data consist of single-turn reasoning prompts covering science, humanities, and logical reasoning, with no multi-turn or agent data.

For each model, the trained policy is evaluated at \high, \med, and \low and compared with its initialization checkpoint evaluated at \high. The initialization checkpoint also serves as the frozen reference policy for the dense anchor. The evaluation suite comprises 24 benchmarks shared by both models across four categories: \textbf{General:} AA-LCR, LongBench-v2, IFEval, and IFBench~\citep{artificialanalysis2025lcr,DBLP:journals/corr/abs-2311-07911,DBLP:conf/nips/PyatkinMGIHDLH25}; \textbf{Reasoning:} HLE, GPQA, AIME26, HMMT-202602, ARC-AGI-1, and ARC-AGI-2~\citep{phan2025lastexam,rein2024gpqa,dekoninck2026matharena,chollet2019arc,DBLP:journals/corr/abs-2505-11831}; \textbf{Code:} SciCode, swe-compass-cd (CC), swe-compass-cu (CC, 0629), Terminal-Bench~2.1, and deploy-full-200 (CC-v3)~\citep{tian2024scicoderesearchcodingbenchmark,merrill2026terminal}; and \textbf{Agent (9):} pev-eval (CC), pev-eval (OC), tau2-retail, tau2-airline, tau2-telecom, and the cross-domain, delivery, instore, and OTA variants of VitaBench~\citep{envloop_pev,yao2024taubenchbenchmarktoolagentuserinteraction,DBLP:journals/corr/abs-2506-07982,DBLP:conf/iclr/HeSHHXGSC26}. Task scores are reported for all 24 benchmarks, and thinking lengths are reported for the 20 benchmarks with available length records. \emph{Relative thinking length} is defined as the mean of the per-benchmark ratios of thinking length to that of the reference policy. Please refer to \cref{app:benchmarks,sec:length-protocol} for details.

\begin{figure}[!htb]
\centering
\includegraphics[width=\linewidth]{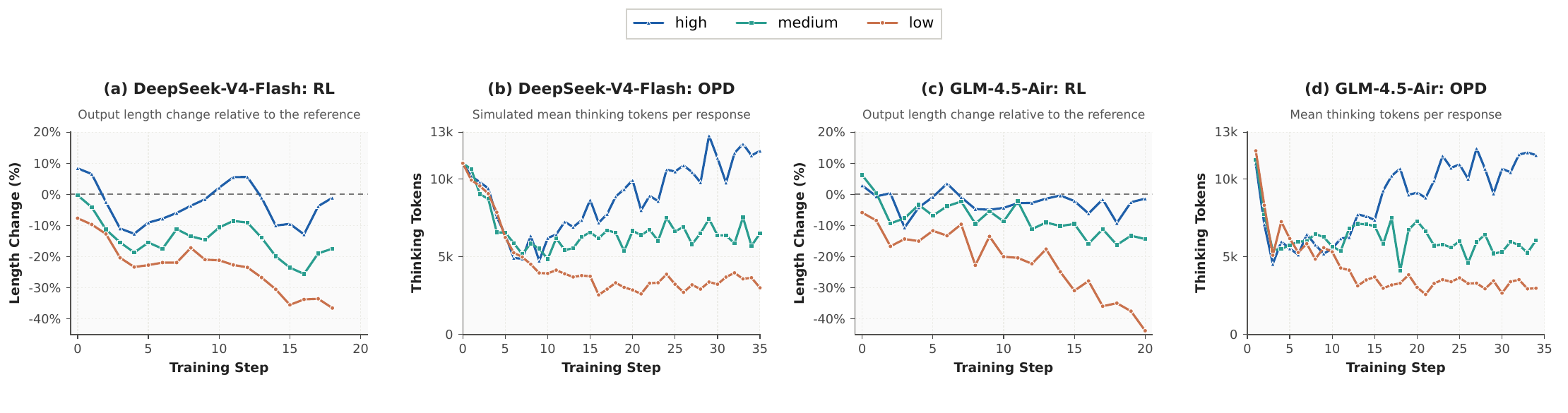}
\caption{\textbf{Thinking length during training.} (a,b)~Prompt-guided dense-anchor RL on DeepSeek-V4-Flash and GLM-4.5-Air: length of each tier relative to the reference policy. (c)~Multi-teacher OPD on GLM-4.5-Air: mean thinking tokens per response for each tier. Values are reconstructed from training plots; please refer to \cref{sec:curve-provenance} for details.}
\label{fig:effort-dynamics}
\end{figure}

\paragraph{Results.}
\textbf{Effort control emerges quickly on general and reasoning tasks.}
The observed RL curves show early tier separation (\cref{fig:effort-dynamics}a,b), without the prolonged plateau of Recipe I. At the evaluated step-20 and step-25 checkpoints, thinking length is strictly ordered as $\high>\med>\low$ on each of the 10 shared general and reasoning benchmarks for both models. In high/medium/low order, relative thinking lengths are $93\%$/$57\%$/$34\%$ for DeepSeek-V4-Flash and $93\%$/$69\%$/$42\%$ for GLM-4.5-Air. On this subset, \high slightly exceeds the reference mean score ($67.22$ vs.\ $66.97$ and $64.85$ vs.\ $64.00$). Shortening has different costs across categories: DeepSeek's general score remains near $70.0$, while its reasoning score falls from $65.25$ to $62.22$ and $58.12$. GLM's reasoning scores are $61.16$, $61.00$, and $57.22$, with the largest loss at \low. Please refer to \cref{app:recipe-lengths} for the per-benchmark numbers.

\begin{figure}[!htb]
\centering
\includegraphics[width=\linewidth]{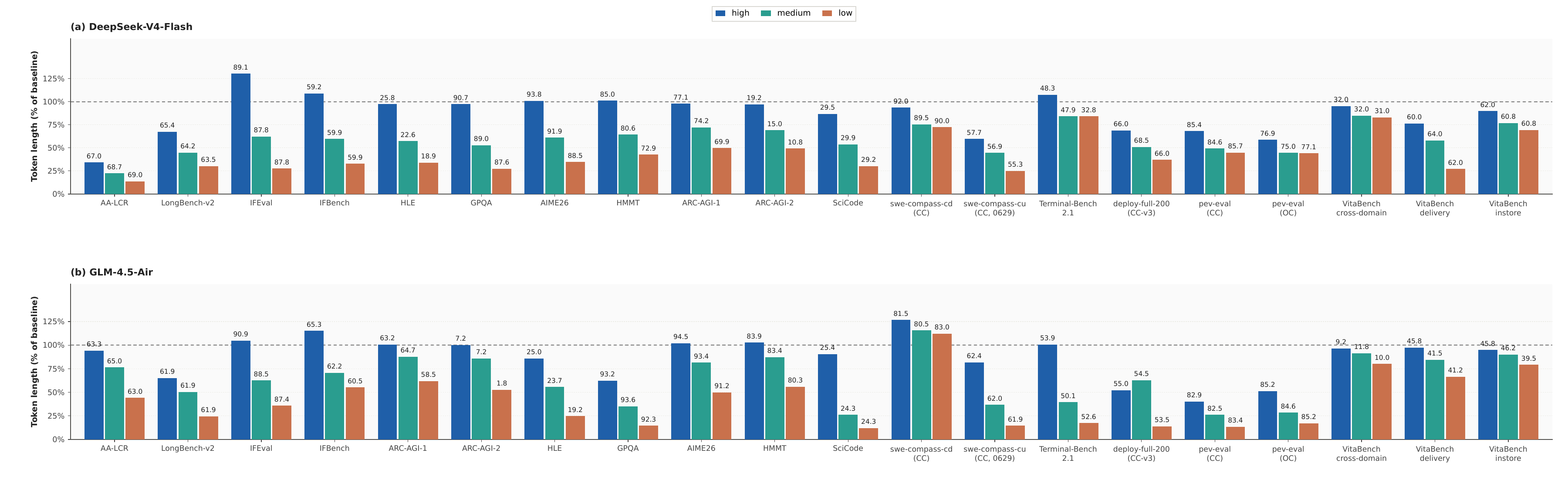}
\caption{\textbf{Thinking length and task score of the prompt-guided policy on each benchmark.} Bars give the thinking length of each tier relative to the reference; the number above a bar is the task score at that tier. The label below each benchmark gives the reference score and its mean thinking tokens. Please refer to \cref{app:recipe-lengths} for the full numbers.}
\label{fig:length-score-bars}
\end{figure}

\begin{figure}[!htb]
\centering
\includegraphics[width=0.86\linewidth]{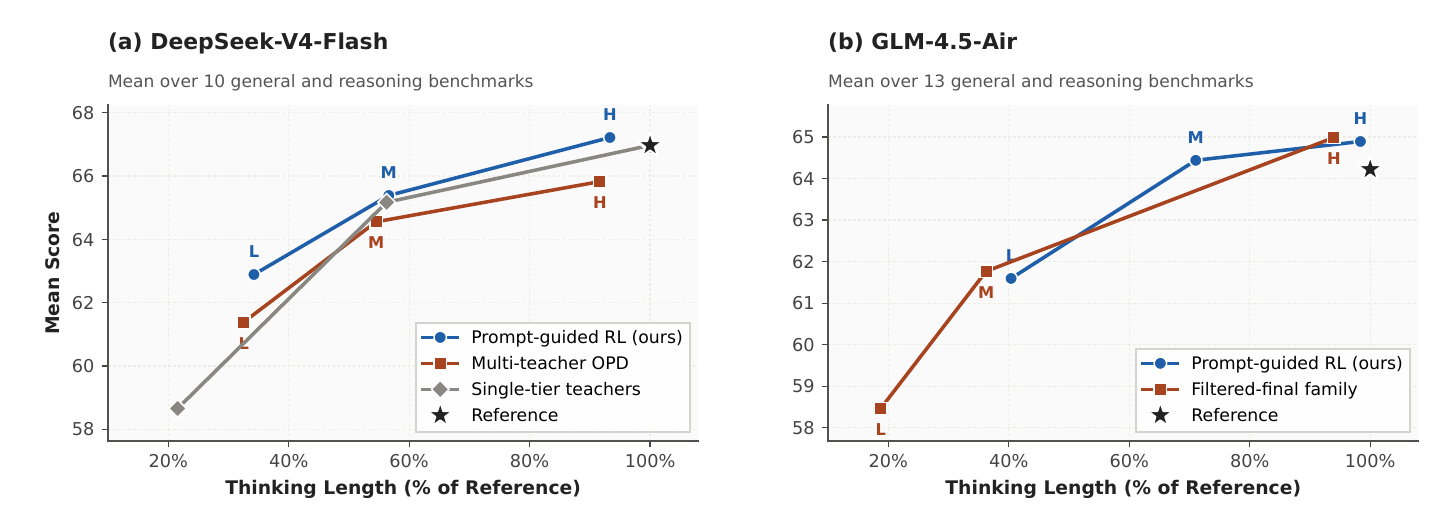}
\caption{\textbf{Score versus thinking length.} Mean score against relative thinking length on the general and reasoning strategies with recorded lengths. Each curve connects the \high (H), \med (M), and \low (L) settings of one checkpoint family; the single-tier teachers are separate checkpoints. The star marks the reference.}
\label{fig:score-length}
\end{figure}

\textbf{At matched length, the recipe is competitive with multi-teacher OPD on general and reasoning tasks.}
On DeepSeek-V4-Flash (\cref{fig:score-length}a), the prompt-guided policy and the OPD student operate at nearly the same relative lengths ($93\%$/$57\%$/$34\%$ vs.\ $92\%$/$55\%$/$33\%$), and the prompt-guided policy scores higher at every tier: $67.22$ vs.\ $65.82$, $65.39$ vs.\ $64.55$, and $62.89$ vs.\ $61.38$. It also matches the single-tier \med teacher ($65.17$ at $56\%$) within one shared policy and without training any teacher. On GLM-4.5-Air (\cref{fig:score-length}b), the families have similar scores at nearby observed lengths: prompt-guided \high scores $64.85$ at $93\%$ versus $64.78$ at $96\%$ for the \high tier of the earlier effort checkpoints; prompt-guided \low scores $61.62$ at $42\%$ versus $61.42$ at $40\%$ for their \med tier.

\textbf{Effort control transfers to multi-turn code and agent tasks.}
Both policies are trained on single-turn prompts only, yet the effort request remains effective on the ten code and agent benchmarks with length records (\cref{fig:length-score-bars}). Thinking length is ordered as $\high \ge \med \ge \low$ on 19 of the 20 model--benchmark pairs. On DeepSeek-V4-Flash, \med and \low use $74\%$ and $57\%$ of the \high length on code benchmarks and $80\%$ and $68\%$ on agent benchmarks; on GLM-4.5-Air, they use $65\%$ and $33\%$ on code and $79\%$ and $60\%$ on agent benchmarks. Under reasoning-only OPD with discrete tags, by contrast, \low still uses $95\%$ and $88\%$ of the \high thinking tokens on code and agent benchmarks (\cref{sec:mech-transfer}). A natural-language instruction states the intended reasoning depth in the input, so its meaning does not have to be learned from multi-turn states. Please refer to \cref{app:recipe-scores} for the task scores by category.

\begin{tcolorbox}[enhanced, colback=phenobg!60, colframe=phenoframe!35, boxrule=0.45pt, arc=1.8mm, borderline west={2.6pt}{0pt}{phenoframe}, fontupper=\small, left=3mm, right=3mm, top=1.6mm, bottom=1.6mm]
\textcolor{phenoframe}{\textbf{Core Recipe Insight (Token-Level Anchoring + Semantic Prompt Priors).}}
\emph{Single-stage effort control requires just two ingredients:} a \textbf{token-level KL anchor on \high} to prevent cross-tier collapse during sparse RL compression, and \textbf{natural-language effort prompts} to seed a format-invariant prior that converges rapidly and transfers zero-shot from single-turn data to multi-turn agents.
\end{tcolorbox}
\section{Discussion}
\label{sec:discussion}

In this section, we answer three questions that readers may naturally raise about our recipes, and through them explain the motivation behind our design. Both recipes place effort training inside RL and constrain it with a token-level anchor, and each question asks whether a simpler or more direct route would reach the same result.

\begin{tcolorbox}[questionsbox]
\begin{itemize}[topsep=0pt,partopsep=0pt,leftmargin=12pt,itemsep=1.5pt]
\item \textbf{Can supervised fine-tuning alone learn effort control?} Effort control can be cast as a supervised problem: collect responses of different lengths, label each with its tier, and fit the policy to them. If supervised fine-tuning (SFT) alone could teach the policy to reason briefly under \low and at length under \high without losing accuracy, the RL stage would be unnecessary (\cref{sec:disc-sft}).
\item \textbf{Can the data mixture alone preserve the \high tier?} \Cref{sec:recipe} studies a constraint added to RL and its combination with tier instructions, but not the data mixture on its own. Sparse Mixed RL transfers tier control across interaction formats, and its only failure is the downward drift of the \high tier. If sampling \high more often could stop this drift, the anchor would be unnecessary (\cref{sec:disc-data}).
\item \textbf{Why not train effort teachers for every domain?} Multi-teacher OPD separates the tiers on every interaction format that its training covers, and its only failure is transfer to formats it does not cover. If effort teachers were trained for every important domain, no transfer would be required, and OPD alone would suffice (\cref{sec:disc-efficiency}).
\end{itemize}
\end{tcolorbox}

\subsection{Can Supervised Fine-Tuning Alone Learn Effort Control?}
\label{sec:disc-sft}

\begin{figure}[!htb]
\centering
\includegraphics[width=\linewidth]{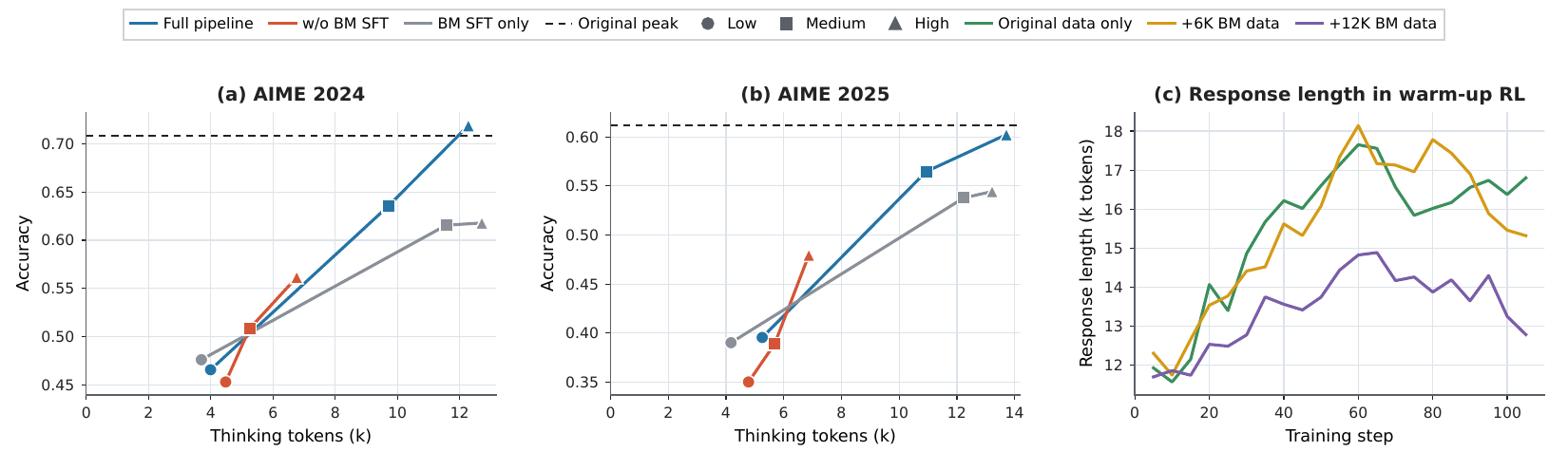}
\caption{\textbf{Budget-mode SFT in ThinkDial.} (a, b)~Accuracy against thinking tokens for the low, medium, and high modes, for the full pipeline, RL without budget-mode (BM) SFT, and BM SFT without RL; the dashed line marks the peak accuracy before effort training. (c)~Response length during warm-up RL with 0, 6K, or 12K BM samples. Redrawn from ThinkDial~\citep{DBLP:journals/corr/abs-2508-18773}.}
\label{fig:disc-sft}
\end{figure}

A natural question is whether effort control requires RL or OPD at all: if demonstrations at each budget are available, the tiers could be fitted directly with SFT. ThinkDial~\citep{DBLP:journals/corr/abs-2508-18773} answers this question with an ablation of its pipeline, which applies budget-mode SFT followed by RL with length rewards (\cref{fig:disc-sft}a,b). With SFT alone, the low mode separates from the others, but the medium and high modes remain close in both thinking length and accuracy. The intended accuracy--length trade-off across the three modes appears only after the RL stage. SFT therefore does not suffice.

Two reasons explain this outcome. The first concerns the data. Lower-effort demonstrations do not occur naturally and have to be constructed, and ThinkDial obtains them by truncating complete reasoning chains. A truncated chain is not a sample from a policy that reasons briefly: the removed reasoning reappears in the answer, and the subsequent RL stage reinforces this pattern until a dedicated penalty is added. SFT maximizes the likelihood of these fixed traces. Its objective contains no term that evaluates the outcome of the policy's own shortened reasoning, whereas a length-penalized reward is computed on the policy's rollouts and acts on its stopping decision directly.

The second reason is that SFT does not remove the coupling between tiers. The tiers share parameters in SFT as they do in RL (\cref{sec:mech-compete}), so supervision on the lower tiers also moves the higher ones. ThinkDial reports this effect: doubling the budget-mode data from 6K to 12K samples shortens the responses of all modes in the subsequent RL stage, from roughly $16$k to below $14$k tokens (\cref{fig:disc-sft}c). SFT thus gives each tier an initial meaning, which is the role natural-language instructions play in Recipe~II without constructed demonstrations (\cref{sec:prompt-opd}), but the trade-off itself is learned from an on-policy signal.

\subsection{Can the Data Mixture Alone Preserve the High Tier?}
\label{sec:disc-data}

\begin{figure}[t]
\centering
\includegraphics[width=\linewidth]{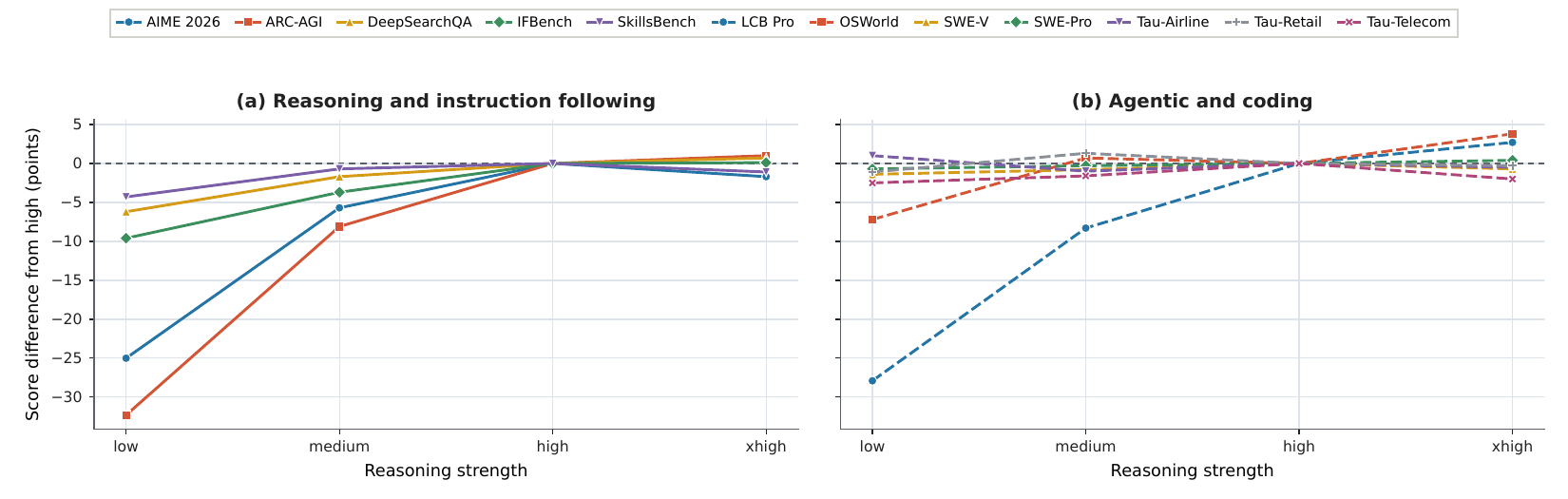}
\caption{\textbf{Scores of Muse Glimmer across its four reasoning strengths}, as differences from the high strength; the shaded band marks differences within three points. (a)~Single-turn reasoning and instruction following. (b)~Agentic coding, customer-service agents, tool use, GUI grounding, long context, and multilingual agents. Replotted from Muse Glimmer~\citep{DBLP:journals/corr/abs-2610-24628}.}
\label{fig:disc-strength}
\end{figure}

\Cref{sec:dense-rl} adds a constraint to RL (Recipe~I) and combines this constraint with effort instructions in the data (Recipe~II). It does not consider the remaining option, which keeps sparse Mixed RL unchanged and adjusts only the data. Muse Glimmer~\citep{DBLP:journals/corr/abs-2610-24628} takes this route. It reports that training on the full spectrum of reasoning strengths shortens the reasoning of the higher strengths. This is the mechanism of \cref{sec:mech-compete}: the highest tiers carry no length penalty and no preservation term, so the length gradient of the lower tiers reaches them through the shared parameters. Its remedy is to change the mixture and route a proportion of the prompts to the higher strengths only. The resulting policy retains its higher strengths, and its lower strengths stay within three points of high on agentic coding, tool-use, GUI grounding, long-context, and multilingual agent benchmarks while losing up to $32$ points on single-turn reasoning (\cref{fig:disc-strength}). Sparse Mixed RL with a tuned mixture thus carries over to interaction formats outside its training data, in agreement with \cref{sec:mech-transfer}.

The limitation of this route is that the outcome depends on the mixture itself. Write the sampling weights of \cref{eq:mixed-gradient} as $w_{\high} = \eta$ and $w_m = (1-\eta)\,\tilde{w}_m$ for $m \in \{\low, \med\}$. The length gradient becomes
\begin{equation}
(1-\eta) \sum_{m \in \{\low,\med\}} \tilde{w}_m \lambda_m \nabla_\theta \mathcal{P}_m ,
\label{eq:mixture-scaling}
\end{equation}
so raising $\eta$ scales the component that shortens \high and the component that compresses \low and \med by the same factor $1-\eta$. At a given amount of lower-tier compression, \high has received the same shortening update for every $\eta$: the mixture changes the time scale of the drift and leaves its direction unchanged. What opposes the drift is the task gradient of \high, whose weight grows with $\eta$. By \cref{prop:grpo-invariance}, this gradient is non-zero only on prompts whose rollouts differ in correctness, and it favors longer reasoning only where shorter reasoning lowers correctness. The value of $\eta$ at which the two forces balance therefore depends on the difficulty of the prompts, the number of tiers, and the penalty weights $\lambda_m$, and it has to be searched again whenever one of them changes. The token-level anchor of \cref{eq:anchor-completes-objective} replaces this search with a preservation term that is active at every state \high visits, independent of the mixture.

\subsection{Why Not Train Effort Teachers for Every Domain?}
\label{sec:disc-efficiency}

\begin{figure}[!htb]
\centering
\includegraphics[width=\linewidth]{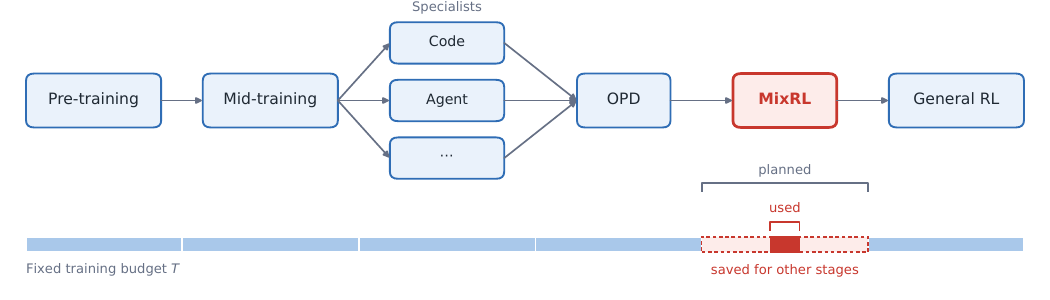}
\caption{\textbf{Effort training within a fixed training budget.} Top: the training pipeline, General RL is an option phase in foundation model training. Bottom: all stages share the budget $C$. The outer bracket marks the time planned for effort training, which per-domain effort teachers would require; the solid segment is the time mixed RL uses, and the dashed segments are saved for the other stages.}
\label{fig:pipeline-budget}
\end{figure}

MOPD separates the tiers on every interaction format that its training covers (\cref{sec:mech-transfer}). Training effort teachers for each important domain and distilling them would therefore also solve the transfer problem. Kimi K3 is a representative example: it trains nine specialists across domains and reasoning-effort tiers and consolidates them with multi-teacher OPD~\citep{DBLP:journals/corr/abs-2607-24653}. Our choice between the two routes is an engineering one. \textbf{A foundation model is not trained with unlimited resources. It is trained to reach a performance target within a fixed budget of compute and time, and all stages of the pipeline share this budget (\cref{fig:pipeline-budget}).}

Scaling laws state how much compute a performance target requires~\citep{DBLP:journals/corr/abs-2001-08361}. When compute is allocated optimally among model size, batch size, and training steps, the test loss follows a power law in the training compute $C$,
\begin{equation}
L(C) = \left( \frac{C_c}{C} \right)^{\alpha_C}, \qquad \alpha_C \approx 0.05 ,
\label{eq:scaling}
\end{equation}
where $C_c$ is a constant. Inverting \cref{eq:scaling} gives the compute that a target loss $L^{\star}$ requires,
\begin{equation}
C(L^{\star}) = C_c \, (L^{\star})^{-1/\alpha_C} .
\label{eq:scaling-inverse}
\end{equation}
The exponent $1/\alpha_C \approx 20$ makes this requirement steep: lowering the target loss by $5\%$ multiplies the required compute by $0.95^{-20} \approx 2.8$.

In practice, the planning runs in the opposite direction. A foundation model is not trained with unlimited resources: the available accelerators and the training schedule fix the budget $C$ before training starts. The expected performance is then combined with this budget to derive what can be trained. \Cref{eq:scaling} gives the loss that the budget can reach, and the same analysis gives the model size that attains it, $N \propto C^{0.73}$~\citep{DBLP:journals/corr/abs-2001-08361}. The budget, not the target, therefore determines the model. All stages of the pipeline draw on this one budget (\cref{fig:pipeline-budget}). If effort training consumes $C_{\mathrm{eff}}$, the remaining stages are planned with $C - C_{\mathrm{eff}}$, and the reachable loss and model size become
\begin{equation}
L = \left( \frac{C_c}{C - C_{\mathrm{eff}}} \right)^{\alpha_C}, \qquad N \propto \big( C - C_{\mathrm{eff}} \big)^{0.73} .
\label{eq:budget}
\end{equation}
Effort training differs from the other stages in that its target is fixed: ordered tiers at retained accuracy. It does not lower $L$, and compute spent beyond the minimum that reaches this target yields no return. Every unit of $C_{\mathrm{eff}}$ is thus taken from the stages whose loss still decreases with compute, and by \cref{eq:scaling-inverse} the performance given up can be bought back only at the exponent $1/\alpha_C$. Under a fixed budget, the efficiency of the effort stage is a design requirement of the method.

The two routes differ in $C_{\mathrm{eff}}$ by a large factor. On a fixed cluster, compute is proportional to training time. With effort teachers for $D$ domains and $|\mathcal{M}|$ tiers, effort training requires on the order of $D \cdot |\mathcal{M}|$ teacher runs followed by distillation, and the runs on agentic domains are paid at the agentic step cost: one training step on agentic data takes several hours because of multi-turn environment interaction and tool execution, whereas one step on single-turn reasoning or general data takes 15--20 minutes. Prompt-guided dense-anchor RL separates the tiers in about 20 single-turn steps (\cref{sec:prompt-opd}), five to seven hours in total, which is comparable to a few steps of a single agentic teacher, and its tier ordering carries over to multi-turn code and agent benchmarks without agentic training data. This is why we pursue recipes that generalize from single-turn data: the smaller $C_{\mathrm{eff}}$ is, the more of the budget remains for the stages that determine $L$ and $N$ in \cref{eq:budget}.

\section{Related Work}
\label{sec:related}

\paragraph{Test-Time Compute Scaling and Reasoning Compression.}
Scaling test-time compute by generating extended chains of thought (CoT) via reinforcement learning with verifiable rewards (RLVR) has substantially advanced LLM reasoning across mathematics, coding, and agentic engineering~\citep{DBLP:journals/nature/GuoYZSWZXZMBZY025, DBLP:journals/corr/abs-2402-03300, DBLP:journals/corr/abs-2501-12599, DBLP:journals/corr/abs-2505-09388, DBLP:journals/corr/abs-2609-23457}. However, RL-trained reasoning models frequently suffer from severe \emph{overthinking}, redundantly re-verifying established conclusions through late backtracking bursts and expending excessive tokens on simple queries~\citep{DBLP:journals/corr/abs-2412-21187, DBLP:journals/tmlr/SuiCWZZYLWZZCH25, DBLP:journals/corr/abs-2605-27965}. To compress reasoning toward a more efficient operating point, existing methods apply supervised fine-tuning or dual-granularity token compression on shortened trajectories~\citep{DBLP:conf/acl/MaWYFW25, DBLP:conf/emnlp/XiaLWLL25, DBLP:journals/corr/abs-2601-20467}, augment RL objectives with sequence- or step-level length penalties and reasoning-shaping schedules~\citep{DBLP:conf/nips/AroraZ25, DBLP:journals/corr/abs-2507-04348, DBLP:journals/corr/abs-2510-09535, DBLP:conf/acl/ChengCFZ26, DBLP:journals/tmlr/HouZJLQAC26, DBLP:conf/acl/YuanXHZGLWZ26, DBLP:journals/corr/abs-2601-06052, DBLP:journals/corr/abs-2602-20945}, or selectively preserve high-value reasoning tokens while pruning routine steps~\citep{DBLP:journals/corr/abs-2609-33970}. Notably, \citet{DBLP:journals/corr/abs-2601-06052} observe that single-policy RL compression trained on a single domain transfers well to out-of-distribution tasks. While these approaches effectively shorten a single policy, they target a fixed efficiency--accuracy trade-off and do not explain why sequence-level RL rewards collapse when jointly training multiple effort tiers within a shared model.

\paragraph{Controllable and Multi-Tier Reasoning Effort.}
Rather than committing to a single compression ratio, recent work seeks to equip a single policy with adjustable reasoning effort along the accuracy--compute Pareto frontier~\citep{DBLP:journals/corr/abs-2507-02076, DBLP:journals/corr/abs-2510-27042, DBLP:journals/corr/abs-2605-16938}. Early approaches framed effort control as a binary switch between thinking and non-thinking modes~\citep{DBLP:conf/emnlp/ZhangLHFL25, DBLP:journals/corr/abs-2505-11896, DBLP:journals/corr/abs-2504-09858, DBLP:journals/corr/abs-2505-09388}, whereas subsequent studies introduced graded control via budget-forcing heuristics~\citep{DBLP:conf/emnlp/MuennighoffYSLFHZLCH25}, length- and budget-conditioned RL or distillation~\citep{DBLP:journals/corr/abs-2503-04697, DBLP:journals/corr/abs-2508-17196, DBLP:journals/corr/abs-2511-01470, DBLP:journals/tmlr/HuangWZSFCF26}, or discrete effort tiers ($\low, \med, \high$)~\citep{DBLP:journals/corr/abs-2508-10925, DBLP:journals/corr/abs-2508-18773, DBLP:journals/corr/abs-2601-08310, DBLP:journals/corr/abs-2606-19348}. To train such multi-tier policies, existing recipes rely either on joint RL with mode-specific length penalties—which triggers severe cross-tier interference and degrades high-effort performance~\citep{DBLP:journals/corr/abs-2508-18773}—or on multi-stage pipelines that first train separate budget teachers and then consolidate them via token-level multi-teacher on-policy distillation (OPD)~\citep{DBLP:conf/iclr/AgarwalVZSGGB24, DBLP:journals/air/FangYCCWLYLGLMRAZWLLZKZ26, DBLP:journals/corr/abs-2602-12125, DBLP:journals/corr/abs-2601-08310, DBLP:journals/corr/abs-2606-30406, DBLP:journals/corr/abs-2601-02780, DBLP:journals/corr/abs-2607-24653, DBLP:journals/corr/abs-2605-10889}. Despite these empirical advances, the underlying mechanism governing when multi-tier effort control succeeds or fails across domains and formats remains uncharacterized. In contrast, we unify sparse RL and dense OPD through supervisory signal density—explaining why sparse Mixed RL collapses across tiers while dense multi-teacher OPD fails zero-shot on unseen multi-turn agentic formats—and show that a single token-level KL anchor on \high combined with natural-language effort prompts achieves both single-stage controllability and zero-shot agentic generalization.

\section{Conclusion}
\label{sec:conclusion}

\paragraph{Conclusion.} This work studies reasoning-effort training and traces the trade-off between multi-tier controllability and cross-format generalization to a single governing principle: supervisory signal density. Sparse sequence-level rewards shift global stopping priors that generalize zero-shot across interaction formats, yet broadcast uniform credit across tokens and collapse shared multi-tier policies. Conversely, dense token-level supervision isolates effort tiers at sparse reflection and stopping transitions, yet binds controllability to the interaction structures visited during training. Guided by this mechanism, we show that multi-stage teacher cascades are unnecessary: anchoring only the \high tier with a token-level KL loss enables single-stage multi-tier RL, and seeding tiers with natural-language effort prompts aligns RL with the base model's instruction-following prior, converging in 20 steps on single-turn data while generalizing zero-shot to unseen multi-turn agents.

\paragraph{Limitations and Future Work.}
Our framework has two primary limitations that point to natural directions for future work. First, the target effort tier ($\low, \med, \high$) is specified externally at inference time and restricted to three discrete modes; learning an intrinsic router that estimates query difficulty beforehand and dynamically adapts continuous or fine-grained compute budgets across turns in long-horizon agentic tasks is a promising next step. Second, while our Prompt-Guided Dense-Anchor RL transfers zero-shot from single-turn prompts to unseen multi-turn agents, its initial length separation relies on the base policy already possessing a sufficiently strong instruction-following prior. Finally, our finding that anchoring only the top-$10\%$ high-entropy branching and stopping tokens recovers full-sequence anchoring suggests a broader principle beyond length control: selective token-level KL anchoring could be applied to resolve cross-objective interference in general post-training, jointly optimizing deep reasoning, safety alignment, and stylistic constraints without degrading frontier capabilities.

\clearpage
\bibliography{reference}
\bibliographystyle{colm2024_conference}

\clearpage
\appendix
% Float layout for the appendix: tables share pages with text and with each other.
\setcounter{topnumber}{4}\setcounter{bottomnumber}{4}\setcounter{totalnumber}{8}
\renewcommand{\topfraction}{0.95}\renewcommand{\bottomfraction}{0.95}
\renewcommand{\textfraction}{0.05}\renewcommand{\floatpagefraction}{0.75}
\makeatletter
\setlength{\@fptop}{0pt}\setlength{\@fpsep}{12pt plus 2pt}\setlength{\@fpbot}{0pt plus 1fil}
\makeatother

\section{Training Data and Evaluation Benchmarks}
\label{app:details}
\label{app:cap-vs-learned}
\label{app:prompts}
\label{app:stats}

\subsection{Training Data by Experiment}
\label{app:train-data}
\label{app:single-compress-details}
\label{app:configs}
\paragraph{Compression and distillation.}
The Qwen3-8B compression run used AIME mathematics and IFEval instruction-following prompts. HMMT and IFBench tested transfer to other benchmarks in the same domains. The process-reward comparison used a mixture of AIME25 and IFEval. In the original GLM compression run, the mixture included general-agent tasks and science reasoning, with IFEval added during training. Its 30,000 base prompts yielded 90,000 effort-conditioned training entries by pairing each prompt with three effort tiers and their respective teachers.

\paragraph{Format transfer and dense-anchor RL.}
The format-transfer experiment trained on single-turn mathematics, science, and instruction following, then evaluated the resulting policies on interactive agent tasks. The comparison with OPD trained on agent data tests whether exposure to interaction during training affects effort control. For Recipe~I, we used reasoning prompts and a subset of the instruction-following data. Recipe~II used single-turn prompts covering humanities, science, and logical reasoning. 

\subsection{Evaluation Benchmarks}
\label{app:benchmarks}
We group the evaluations by the task given to the model. \Cref{tab:benchmark-overview} lists the benchmarks of the two evaluation suites with their scoring method; per-benchmark scores are reported in \cref{tab:d-export,tab:g-export,tab:aggregate}.

\begin{table}[!htb]
\centering\footnotesize
\setlength{\tabcolsep}{6pt}
\renewcommand{\arraystretch}{1.02}
\begin{tabular}{@{}c l l c c@{}}
\toprule
 & \textbf{Benchmark} & \textbf{Scoring} & \textbf{DeepSeek-V4-Flash} & \textbf{GLM-4.5-Air} \\
\midrule
\multirow{7}{*}{\rotatebox[origin=c]{90}{\textbf{General}}} & AA-LCR & LLM judge (Qwen3-235B) & $\checkmark$ & $\checkmark$ \\
 & LongBench-v2 (128k) & Automatic & $\checkmark$ & $\checkmark$ \\
 & MRCR (32k) & Automatic &  & $\checkmark$ \\
 & IFEval & Automatic & $\checkmark$ & $\checkmark$ \\
 & IFBench & Automatic & $\checkmark$ & $\checkmark$ \\
 & AdvancedIF & LLM judge (Gemini 3.1 Flash-Lite) &  & $\checkmark$ \\
 & ChineseSimpleQA & LLM judge (Gemini 3.1 Flash-Lite) &  & $\checkmark$ \\
\midrule
\multirow{6}{*}{\rotatebox[origin=c]{90}{\textbf{Reasoning}}} & AIME26 & Automatic & $\checkmark$ & $\checkmark$ \\
 & HMMT-2602 & Automatic & $\checkmark$ & $\checkmark$ \\
 & GPQA-Diamond & Automatic & $\checkmark$ & $\checkmark$ \\
 & HLE & LLM judge (Gemini 3.1 Flash-Lite) & $\checkmark$ & $\checkmark$ \\
 & ARC-AGI-1 & Automatic & $\checkmark$ & $\checkmark$ \\
 & ARC-AGI-2 & Automatic & $\checkmark$ & $\checkmark$ \\
\midrule
\multirow{6}{*}{\rotatebox[origin=c]{90}{\textbf{Code}}} & SciCode & Automatic & $\checkmark$ & $\checkmark$ \\
 & SWE-Compass (CD) & Automatic & $\checkmark$ & $\checkmark$ \\
 & SWE-Compass (CU) & LLM judge (GPT-5.2) & $\checkmark$ & $\checkmark$ \\
 & Terminal-Bench 2.1 & Automatic & $\checkmark$ & $\checkmark$ \\
 & CCBench Deploy & Automatic & $\checkmark$ & $\checkmark$ \\
 & CCBench Understand & LLM judge (GPT-5.2) &  & $\checkmark$ \\
\midrule
\multirow{14}{*}{\rotatebox[origin=c]{90}{\textbf{Agent}}} & Pev-Bench (CC) & LLM judge (GPT-5.4) & $\checkmark$ & $\checkmark$ \\
 & Pev-Bench (OC) & LLM judge (GPT-5.4) & $\checkmark$ & $\checkmark$ \\
 & $\tau^2$-Bench Retail & Automatic & $\checkmark$ & $\checkmark$ \\
 & $\tau^2$-Bench Airline & Automatic & $\checkmark$ & $\checkmark$ \\
 & $\tau^2$-Bench Telecom & Automatic & $\checkmark$ & $\checkmark$ \\
 & VitaBench Cross-domain & LLM judge (GPT-5) & $\checkmark$ & $\checkmark$ \\
 & VitaBench Delivery & LLM judge (GPT-5) & $\checkmark$ & $\checkmark$ \\
 & VitaBench In-store & LLM judge (GPT-5) & $\checkmark$ & $\checkmark$ \\
 & VitaBench OTA & LLM judge (GPT-5) & $\checkmark$ & $\checkmark$ \\
 & DSBench DataModeling & Automatic &  & $\checkmark$ \\
 & DSBench DataAnalysis & Automatic &  & $\checkmark$ \\
 & SpreadsheetBench Verified & Automatic &  & $\checkmark$ \\
 & OfficeQA & Automatic &  & $\checkmark$ \\
 & OfficeQA Pro & Automatic &  & $\checkmark$ \\
\bottomrule
\end{tabular}
\caption{\textbf{Benchmarks in the two evaluation suites.} The DeepSeek-V4-Flash suite contains 24 benchmarks and the GLM-4.5-Air suite contains 33; the 24 shared benchmarks form the common suite. Automatic scoring uses the benchmark's own checker or environment verifier; the remaining benchmarks are scored by the listed LLM judge.}
\label{tab:benchmark-overview}
\end{table}

\subsubsection{Mathematical, Scientific, and Abstract Reasoning}
\paragraph{AIME \citep{dekoninck2026matharena}.}The model solves competition mathematics problems and returns an integer answer. The paired suite uses the English 2026 edition. \Cref{tab:aggregate} reports the 2025 and 2026 editions, each with separate scores for tool-assisted and tool-free evaluation. Answers range from 000 to 999. Substantial differences in the length of two derivations can therefore lead to the same scored answer.

\paragraph{HMMT \citep{dekoninck2026matharena}.} Problems from the Harvard--MIT Mathematics Tournament require a final mathematical answer. The paired suite uses the English February 2026 contest. \Cref{tab:aggregate} also includes the February and November 2025 contests and identifies the tool-assisted variants separately.

\paragraph{GPQA \citep{rein2024gpqa}.} Graduate-level multiple-choice questions test reasoning in biology, chemistry, and physics. The paired comparison uses the English variant. The earlier OPD evaluation in \cref{tab:aggregate} uses the Diamond subset. The distractors are designed to be plausible to readers with general scientific knowledge, so solving a question often requires applying a domain-specific principle.

\paragraph{Humanity's Last Exam (HLE) \citep{phan2025lastexam}.} The model answers expert-written questions across academic disciplines. We use the judged variant, in which an evaluator assesses the final answer against the reference.

\paragraph{MMLU-Pro \citep{DBLP:conf/nips/WangMZNCGRAHJLK24}.} The model selects an answer to a multiple-choice question drawn from a range of academic and professional subjects. The questions emphasize using subject knowledge to reason through plausible alternatives.

\paragraph{BBEH \citep{DBLP:conf/acl/KazemiFBPAMJAJC25}.} BIG-Bench Extra Hard comprises demanding reasoning tasks with task-specific instructions. The model must apply the rules in the prompt to derive the answer.

\paragraph{ARC-AGI-1 \citep{chollet2019arc}.} Each puzzle gives examples of input and output grids. The model infers the transformation illustrated by those pairs and predicts the output grid for a new input. For example, a rule may move objects of one color while preserving their shapes. The output must specify the entire grid, including the cells left unchanged.

\paragraph{ARC-AGI-2 \citep{DBLP:journals/corr/abs-2505-11831}.} The second suite uses the same grid-prediction format, with puzzles that require interpreting symbols in context or composing multiple transformations. Each puzzle supplies its own demonstrations.

\paragraph{ChineseSimpleQA \citep{DBLP:conf/acl/HeLLTWHBGHZLSZS25}.} Short Chinese-language questions ask for factual answers. The task tests whether the model can supply the requested fact precisely.

\subsubsection{Long-Context Reasoning and Retrieval}
\paragraph{AA-LCR \citep{artificialanalysis2025lcr}.} The model reads a collection of long documents and answers a question that draws on evidence across them. This tests whether the model can locate relevant passages and reason about them together. Questions can require combining evidence from several financial reports, research papers, or legal documents before drawing a conclusion.

\paragraph{LongBench-v2 \citep{DBLP:conf/acl/BaiTZ0WLCX0D0L25}.} Given a long context and a multiple-choice question, the model selects the answer supported by the context. We use the 128k variant. The answer choices can differ on details spread throughout the input, so locating a relevant passage is only the first step in selecting an answer.

\paragraph{MRCR \citep{openai2025mrcr}.} A long conversation contains repeated, similar requests. The model must retrieve the response associated with the specified occurrence. We use the 32k context variant. Multiple requests in the same history share their wording or topic. The position specified in the query determines which earlier response the model must reproduce.

\paragraph{RULER \citep{hsieh2024ruler}.} RULER uses synthetic tasks to test retrieval, state tracking, and aggregation over long inputs. The OPD suite uses the 64k setting.

\paragraph{LongProc \citep{DBLP:journals/corr/abs-2501-05414}.} This benchmark evaluates long-context procedural tasks. The model follows the instructions in the input to produce the requested output.

\subsubsection{Instruction Following and Dialogue}
\paragraph{IFEval \citep{DBLP:journals/corr/abs-2311-07911}.} Each prompt combines a writing task with explicit instructions, such as including a keyword or using a prescribed format. Programmatic checks determine which instructions the response satisfies. For example, the same response may have to contain a specified phrase and obey a length limit. These conditions can be evaluated on the generated text itself.

\paragraph{IFBench \citep{DBLP:conf/nips/PyatkinMGIHDLH25}.} The model generates text under verifiable constraints on properties such as word position, counts, or repetition. A response can answer the underlying question correctly and still fail a constraint. For instance, requiring a word at a particular position tests whether the model can maintain a constraint while composing the rest of the response.

\paragraph{AdvancedIF \citep{DBLP:conf/acl/HeLZLMKXWPLBPQF26}.} This judged instruction-following evaluation measures compliance with a prompt's requirements. The GLM results use the Gemini 3.1 Flash Lite judge.

\paragraph{MultiChallenge \citep{DBLP:conf/acl/DeshpandeSMJHLK25}.} Multi-turn conversations require the model to carry earlier instructions, facts, or corrections into a later response. The final request can depend on information introduced several turns earlier.

\paragraph{SuperCLUE \citep{DBLP:journals/corr/abs-2307-15020}.} This Chinese-language evaluation probes the model's ability to respond to user prompts. The OPD suite includes its score in the instruction-following and dialogue block.

\subsubsection{Programming and Software Engineering}
\paragraph{SciCode \citep{tian2024scicoderesearchcodingbenchmark}.} The model writes Python code for scientific computing problems designed by domain experts. Problems are divided into subproblems whose implementations can be checked against unit tests. A single problem may ask for several numerical routines that together implement a scientific calculation. The test inputs probe whether the generated routines reproduce the required numerical behavior.

\paragraph{LiveCodeBench \citep{jain2024livecodebenchholisticcontaminationfree}.} LiveCodeBench evaluates code generation on competitive programming problems. Test cases check whether its output satisfies the problem specification. Each problem specifies an input format, an output format, and constraints. A solution must implement a general algorithm that also handles cases beyond the examples in the prompt.

\paragraph{SWE-V (R2E-Gym) \citep{jimenez2024swebench}.} In the SWE-bench task formulation, the agent receives a repository and an issue description and produces a patch. The OPD suite uses the R2E-Gym environment for this software-engineering evaluation.

\paragraph{SWE-Compass \citep{DBLP:journals/corr/abs-2511-05459}.} This agentic software-engineering evaluation uses the CD task set under the CC scaffold. Its score contributes one entry to the code category mean. We also evaluate the CU task set under the CC scaffold and report its 0629 revision as a separate strategy.

\paragraph{Terminal-Bench 2.1 \citep{merrill2026terminal}.} An agent receives a task and a terminal environment in which to carry it out. Task-specific checks evaluate the resulting environment. The paired suite uses version 2.1, while \cref{tab:aggregate} reports Terminal-Bench~2. An agent can inspect files, execute commands, and use the resulting output to choose its next action. Success depends on the files or environment state left at the end of the interaction.

\paragraph{CCBench \citep{DBLP:conf/iiswc/FanWLLTLLSYYLCZTT26}.} This deployment evaluation runs under the CC-v3 scaffold. The code results in \cref{tab:g-export,tab:d-export} report this configuration as deploy-full-200 (CC-v3). In addition, the code-understanding strategy is evaluated with the CC-v1 scaffold and reported under that revision in the GLM score table.

\subsubsection{Data Analysis and Document Tasks}
\paragraph{DSBench DataModeling \citep{DBLP:journals/corr/abs-2511-05459}.} Given a dataset and a modeling objective, the agent develops a data-modeling solution. Our evaluation provides a terminal environment. The model must turn the supplied data into a usable modeling workflow, including preparing inputs and producing predictions for the stated target.

\paragraph{DSBench DataAnalysis \citep{DBLP:journals/corr/abs-2511-05459}.} The agent uses a terminal to inspect data and answer an analysis question. The GLM suite reports this task separately from DSBench DataModeling. The requested result may require inspecting columns, computing a summary, and interpreting the result in light of the question.

\paragraph{SpreadsheetBench Verified \citep{ma2024spreadsheetbench}.} The agent modifies a spreadsheet according to a natural-language instruction. The GLM suite evaluates the Verified variant in a terminal, and the earlier OPD table reports SpreadsheetBench under its original label. A task starts from an existing workbook. The relevant output is the modified workbook, so the model must identify the intended cells or ranges and encode the requested changes there.

\paragraph{OfficeQA \citep{officeqa}.} The model answers questions using evidence in office documents. The GLM evaluation gives the agent terminal access to work with the documents.

\paragraph{OfficeQA Pro \citep{DBLP:journals/corr/abs-2603-08655}.} The Pro task set provides a second document question-answering evaluation under the terminal setup. Its results appear separately from OfficeQA.

\subsubsection{Tool Use and Information Seeking}
\paragraph{ACEBench \citep{DBLP:conf/emnlp/ChenHLHZYLHLWL25}.} The model interprets a user request and supplies the corresponding tool calls. The OPD evaluation reports separate Chinese and English results. The tool name and the arguments express different parts of the decision. The model must select an operation and translate details from the request into values accepted by that operation.

\paragraph{BFCL-v4 (web) \cite{patil2025bfcl}.} The Berkeley Function Calling Leaderboard evaluates whether a model can use functions to complete a request. We report the web variant of version~4.

\paragraph{MCP-v1 \citep{DBLP:conf/iclr/WangCPBWLDRSBS26}.} This evaluation exercises tool use through the Model Context Protocol. The OPD suite reports the v1 configuration.

\paragraph{DeepSearchQA \citep{DBLP:journals/corr/abs-2601-20975}.} Open-ended questions require research across sources to assemble a complete answer. Some questions require identifying multiple items that meet the stated criteria. For questions asking for a set, the model must collect candidates across sources and check each against the inclusion criteria. Missing a qualifying item changes the completeness of the answer.

\subsubsection{Interactive Service and Agent Tasks}
\paragraph{$\tau^2$-Bench Retail \citep{yao2024taubenchbenchmarktoolagentuserinteraction}.} The $\tau$-bench framework pairs an assistant with a simulated user and domain-specific tools. In the retail domain of $\tau^2$-Bench, the agent assists a customer with requests concerning orders, returns, or exchanges. It must use the available tools and follow store policy. We use the 20260319 revision. A return request, for example, requires identifying the relevant order, checking which changes are permitted, and completing the requested transaction.

\paragraph{$\tau^2$-Bench Airline \citep{DBLP:journals/corr/abs-2506-07982}.} The agent handles airline service requests, such as changing a reservation, while observing booking and fare rules. We use the 20260303 revision. A conversation may require the agent to identify a booking, establish which itinerary changes the customer wants, and use reservation tools to implement an eligible change.

\paragraph{$\tau^2$-Bench Telecom \citep{DBLP:journals/corr/abs-2506-07982}.} The agent helps a simulated customer resolve a telecommunications issue. Tasks can require actions on both the agent and customer sides. We use the 20260319 revision. This creates a dependency across turns. A tool observation or a diagnostic step performed by the customer can determine which action the agent should take next.

\paragraph{VitaBench Cross-domain \citep{DBLP:conf/iclr/HeSHHXGSC26}.} A simulated user presents a request spanning service domains. The agent uses tools from the relevant domains over the course of the interaction.

\paragraph{VitaBench Delivery \citep{DBLP:conf/iclr/HeSHHXGSC26}.} Delivery scenarios place the agent in a conversation about a delivery service request. The agent must gather the necessary details and act through the service tools.

\paragraph{VitaBench In-store \citep{DBLP:conf/iclr/HeSHHXGSC26}.} The simulated user's request concerns a service at a physical venue. The model uses the in-store tools to carry out the request across turns.

\paragraph{VitaBench OTA \citep{DBLP:conf/iclr/HeSHHXGSC26}.} This domain covers travel-service interactions. The agent responds to the user's requirements using the tools available in the OTA environment.

\paragraph{Pev-Bench (CC) \citep{envloop_pev}.} The CC variant evaluates the policy through an external agent scaffold. We report the score obtained with CC as one agent strategy.

\paragraph{Pev-Bench (OC) \citep{envloop_pev}.} The OC variant runs pev-eval through a second external scaffold. Reporting this score separately preserves the distinction between the two interaction settings.

\subsection{Evaluation Protocol}
\label{app:eval-protocol}
\label{app:matched-metrics}
\paragraph{Thinking length and truncation.}
For a response $y$, thinking length $\Lth(y)$ counts the generated tokens preceding the first end-of-thinking marker. A trace with no marker before truncation is counted as censored, and for the bound in \cref{prop:opd-stopping} we set $\Lth(y)=|y|$. We report two ratios with different denominators: the compression ratio relative to a reference checkpoint, $1-\bar L_c/\bar L_0$, and the cross-tier ratio $\bar L_m/\bar L_{\high}$. On agent tasks, the number of environment turns and the thinking tokens per turn are reported separately (\cref{tab:agent-turns}).

\paragraph{Matched comparisons.}
Each comparison holds fixed the prompt set, benchmark variant, sampling configuration, tools, judge, and evaluation cap. To separate learned stopping from forced truncation, the compression checkpoints are evaluated under a relaxed 32k cap (\cref{sec:pheno-compress}).

\subsection{Score Aggregation and Checkpoint Identification}
\label{sec:export-protocol}
The evaluations use the 24 benchmarks shared by both evaluations; GLM has nine additional benchmarks that are excluded from this comparison. Category sizes in the common set are 4 general, 6 reasoning, 5 code, and 9 agent strategies. Total scores average individual strategy scores; small differences from recomputing the displayed values reflect rounding. Column keys in \cref{tab:g-export,tab:d-export} are defined in the table captions. The reference is itself a multi-teacher OPD checkpoint. 

\subsection{Curve Reconstruction}
\label{sec:curve-provenance}
The panels of \cref{fig:effort-dynamics} are reconstructed from training plots by pixel-color digitization without smoothing. Panels (a) and (b) show prompt-guided RL on DeepSeek-V4-Flash and GLM-4.5-Air, respectively, as percentage changes in output length relative to the reference; the GLM source ratios are converted to percentage changes. Panel (c) shows mean thinking tokens for the GLM multi-teacher OPD run.

\subsection{Length Statistics}
\label{sec:length-protocol}
Thinking lengths in \cref{tab:d-len,tab:g-len,tab:d-len-full,tab:g-len-full} are the mean number of thinking tokens per response reported by the evaluation platform for the same evaluation task as the paired score. 

\section{Supplementary Experimental Results}
\label{sec:export-details}

\subsection{Single-Policy Compression across Training Caps}
\label{app:compression-ladder}
\Cref{tab:qwen-ladder,tab:ladder} give the checkpoint evaluations behind \cref{fig:single_compress} in \cref{sec:pheno-compress}. Each entry pairs task accuracy with mean thinking tokens for the checkpoint at the end of a training stage, evaluated under a relaxed 32k cap, so a shorter trace reflects the stopping behavior of the policy and not the decoding limit.

On GLM-4.5-Air, the thinking length on AIME25 falls from $15.4$k to $2.4$k tokens across the stages while accuracy moves from $92.4$ to $90.6$; HMMT follows the same pattern ($17.5$k to $2.5$k, $89.2$ to $86.7$), and GPQA improves from $81.7$ to $89.9$ as its length falls from $6.2$k to $1.1$k. The held-out benchmarks shorten as well: LongBench-v2 and IFBench peak at intermediate stages above the initial checkpoint ($63.1$ and $81.3$ against $58.5$ and $77.5$). Qwen3-8B tolerates less compression: accuracy on AIME25 is preserved down to the 16k stage ($86.3$ at $6.0$k tokens, against $86.5$ at $9.2$k initially) and declines to $76.7$ at the 4k stage, and held-out HMMT declines from $32.0$ to $21.0$. Its instruction-following scores peak at the 8k stage (IFEval $94.8$, IFBench $46.8$) with less than half of the initial thinking length.
\begin{table}[!htb]
\centering\footnotesize
\setlength{\tabcolsep}{4.5pt}
\caption{Staged-cap RL compression ladder on Qwen3-8B: accuracy / mean thinking tokens per stage checkpoint under a relaxed 32k evaluation cap. The lower block denotes held-out out-of-distribution (OOD) benchmarks.}
\label{tab:qwen-ladder}
\begin{tabular}{@{}lcccc@{}}
\toprule
Benchmark & Init / $32\text{k}$ (0) & $16\text{k}$ (100) & $8\text{k}$ (200) & $4\text{k}$ (300) \\
\midrule
AIME25 (ID)     & \win{86.5} / 9.2k & 86.3 / 6.0k & 84.2 / 4.3k & 76.7 / 2.6k \\
IFEval (ID)     & 91.8 / 2.4k & 92.6 / 2.0k & \win{94.8} / 1.1k & 93.5 / 0.6k \\
\midrule
HMMT (OOD)      & \win{32.0} / 11.4k & 31.3 / 7.7k & 26.7 / 4.8k & 21.0 / 2.5k \\
IFBench (OOD)   & 45.8 / 3.8k & 46.2 / 3.5k & \win{46.8} / 1.7k & 43.2 / 0.8k \\
\bottomrule
\end{tabular}
\end{table}

\begin{table}[!htb]
\centering\footnotesize
\setlength{\tabcolsep}{2.2pt}
\caption{Staged-cap RL compression ladder on the base model: accuracy / mean thinking tokens per checkpoint, labeled by active cap (training step). The lower block denotes held-out benchmarks.}
\label{tab:ladder}
\begin{tabular}{@{}lccccccc@{}}
\toprule
Cap (step) & none (0) & 16k (70) & 14k (140) & 12k (210) & 10k (245) & 8k (260) & 6k (315) \\
\midrule
AIME25 & 92.4 / 15.4k & 90.6 / 6.7k & 91.5 / 5.6k & 92.4 / 4.9k & \win{93.0} / 4.6k & 89.6 / 3.2k & 90.6 / 2.4k \\
HMMT   & \win{89.2} / 17.5k & 86.5 / 6.9k & 88.5 / 6.2k & 87.3 / 5.1k & 88.3 / 5.1k & 85.2 / 3.4k & 86.7 / 2.5k \\
GPQA   & 81.7 / 6.2k & 90.9 / 2.8k & 91.0 / 2.6k & \win{93.2} / 2.5k & \win{93.2} / 2.4k & 91.7 / 1.5k & 89.9 / 1.1k \\
\midrule
ChineseSimpleQA & 58.4 / 2.5k & 58.5 / 2.3k & 59.4 / 740 & 59.6 / 814 & 59.8 / 587 & \win{61.1} / 1.4k & 59.6 / 1.0k \\
LongBench-v2 & 58.5 / 2.5k & 60.8 / 2.0k & 60.4 / 1.5k & 62.7 / 1.5k & \win{63.1} / 1.4k & 61.1 / 932 & 58.9 / 689 \\
IFBench & 77.5 / 3.2k & 78.6 / 3.0k & 78.9 / 2.9k & 79.6 / 2.9k & 80.3 / 3.7k & \win{81.3} / 2.5k & 79.2 / 1.7k \\
SuperCLUE & 57.1 / 5.5k & \win{58.5} / 5.1k & 58.0 / 5.2k & 53.5 / 4.8k & 57.7 / 4.7k & 51.8 / 3.5k & 55.2 / 3.2k \\
\bottomrule
\end{tabular}
\end{table}

\subsection{Effort Separation under Multi-Teacher OPD}
\label{app:opd-separation}
For the GLM experiment in \cref{sec:pheno-opd}, \cref{tab:aggregate} reports the task scores of each tier at step~15, followed by the suite-level score and mean thinking length at steps~15 and~20. At step~15, mean thinking length ranges from $3{,}771$ tokens at \low to $9{,}012$ at \high, $60.8\%$ and $6.2\%$ below the reference, and the corresponding suite scores are $66.4$ and $71.0$.

The benchmark-level scores show where the tiers differ. The \high student matches or exceeds the reference on most benchmarks, including the mathematics mean ($88.3$ vs.\ $88.0$) and the agentic mean ($73.1$ vs.\ $72.6$). The cost of lower effort concentrates on long-horizon reasoning: the mathematics mean falls to $83.4$ at \med and $80.3$ at \low, and ARC-AGI-1 falls from $62.8$ to $50.4$ and $44.4$. Instruction-following and agentic scores change little across tiers, with agentic means of $72.0$ and $71.2$ at \med and \low.
\begin{table}[!htb]
\centering\scriptsize
\setlength{\tabcolsep}{3.8pt}
\renewcommand{\arraystretch}{1.05}
\newcommand{\thdr}[2]{\begin{tabular}[c]{@{}c@{}}\textbf{#1}\\[-1.5pt]\textbf{#2}\end{tabular}}
\resizebox{\linewidth}{!}{%
\begin{tabular}{@{}lcccccccccc@{}}
\toprule
 & \multicolumn{10}{c}{\textit{\textbf{Competition Mathematics}}} \\
\cmidrule(l{2pt}){2-11}
\textbf{Model / Tier} & \textbf{AIME25} & \thdr{AIME25}{(tool)} & \textbf{AIME26} & \thdr{AIME26}{(tool)} & \textbf{HMMT25-Feb} & \thdr{HMMT25-Feb}{(tool)} & \textbf{HMMT25-Nov} & \thdr{HMMT25-Nov}{(tool)} & \textbf{HMMT26-Feb} & \thdr{HMMT26-Feb}{(tool)} \\
\midrule
Reference & 92.2 & 97.7 & 92.0 & 96.7 & 88.8 & 91.9 & 84.2 & 94.0 & 74.7 & 77.5 \\
Student \high   & \win{92.9} & \win{96.7} & \win{92.3} & \win{96.7} & \win{89.6} & \win{93.0} & \win{84.1} & \win{92.4} & \win{76.0} & \win{78.8} \\
Student \med    & 89.9 & \win{96.7} & 86.7 & 94.9 & 85.8 & 91.0 & 73.1 & 88.4 & 63.5 & 73.5 \\
Student \low    & 86.6 & 94.9 & 81.7 & 93.8 & 80.8 & 90.4 & 67.5 & 86.5 & 59.6 & 71.3 \\
\midrule
 & \multicolumn{6}{c}{\textit{\textbf{Science \& General Reasoning}}} & \multicolumn{4}{c}{\textit{\textbf{Long-Context \& Retrieval (I)}}} \\
\cmidrule(lr){2-7}\cmidrule(l){8-11}
\textbf{Model / Tier} & \thdr{GPQA-}{Diamond} & \textbf{MMLU-Pro} & \textbf{BBEH} & \thdr{HLE}{(judged)} & \textbf{ARC-AGI-1} & \textbf{ARC-AGI-2} & \textbf{Ruler-64k} & \textbf{LongProc} & \thdr{LongBench-}{v2} & \textbf{MRCR-32k} \\
\midrule
Reference & 81.8 & 78.3 & 68.7 & 17.5 & 57.8 & 4.8 & 94.4 & 78.9 & 60.0 & 56.5 \\
Student \high   & \win{87.9} & \win{79.9} & 68.9 & \win{17.6} & \win{62.8} & \win{5.4} & 94.2 & \win{77.4} & \win{61.2} & 82.1 \\
Student \med    & 86.7 & 78.9 & \win{70.1} & 13.1 & 50.4 & 3.0 & 95.2 & 70.4 & 60.4 & \win{83.4} \\
Student \low    & 86.9 & 79.1 & 64.1 & 11.7 & 44.4 & 1.8 & \win{95.3} & 68.0 & 59.6 & 83.1 \\
\midrule
 & \multicolumn{1}{c}{\textit{\textbf{LC (II)}}} & \multicolumn{7}{c}{\textit{\textbf{Instruction Following \& Dialogue}}} & \multicolumn{2}{c}{\textit{\textbf{Coding \& Agentic (I)}}} \\
\cmidrule(lr){2-2}\cmidrule(lr){3-9}\cmidrule(l){10-11}
\textbf{Model / Tier} & \textbf{AA-LCR} & \textbf{IFEval} & \textbf{IFBench} & \textbf{AdvancedIF} & \thdr{Multi-}{Challenge} & \textbf{SuperCLUE} & \thdr{Arena}{(1126)} & \thdr{Arena}{(0228)} & \thdr{$\tau^2$-}{airline} & \thdr{$\tau^2$-}{retail} \\
\midrule
Reference & 51.0 & 94.1 & 78.9 & 51.0 & 57.0 & 57.7 & 40.0 & 43.6 & 74.4 & 85.8 \\
Student \high   & 50.0 & \win{94.6} & 78.2 & \win{51.6} & \win{58.0} & \win{58.3} & 39.9 & \win{43.0} & \win{76.8} & \win{87.5} \\
Student \med    & 44.3 & \win{94.6} & \win{82.3} & 50.7 & 56.6 & 55.2 & \win{41.7} & 42.9 & \win{76.8} & 85.4 \\
Student \low    & \win{50.3} & 92.8 & 78.6 & 50.5 & 54.3 & 51.8 & 40.7 & 42.6 & 72.0 & 86.5 \\
\midrule
 & \multicolumn{10}{c}{\textit{\textbf{Coding \& Multi-Turn Agentic (II)}}} \\
\cmidrule(l{2pt}){2-11}
\textbf{Model / Tier} & \thdr{$\tau^2$-}{telecom} & \thdr{ACEBench}{(zh)} & \thdr{ACEBench}{(en)} & \thdr{Terminal-}{Bench 2} & \thdr{BFCL-v4}{(web)} & \textbf{MCP-v1} & \thdr{DeepSearch}{QA} & \thdr{SWE-V}{(R2E-Gym)} & \thdr{LiveCode}{Bench} & \thdr{Spreadsheet}{Bench} \\
\midrule
Reference & 93.0 & 80.7 & 73.4 & 49.8 & 57.7 & 61.5 & 75.9 & 67.8 & 78.6 & -- \\
Student \high   & 94.2 & \win{81.6} & 73.6 & \win{49.1} & 60.1 & \win{60.8} & \win{74.7} & \win{68.2} & \win{77.5} & 48.5 \\
Student \med    & 94.6 & 80.8 & 74.8 & 47.6 & \win{60.2} & 59.4 & 73.8 & 67.0 & 75.3 & 47.3 \\
Student \low    & \win{95.4} & 80.8 & \win{76.4} & 43.1 & 58.8 & 60.0 & 74.1 & 67.8 & 74.4 & \win{54.8} \\
\midrule
 & \multicolumn{10}{c}{\textit{\textbf{33-Benchmark Suite Macro Summary (Steps 15 \& 20)}}} \\
\cmidrule(l{2pt}){2-11}
\textbf{Model / Tier} & \textbf{Acc$_{15}$} & \textbf{$\Delta$Acc$_{15}$} & \textbf{Tok$_{15}$} & \textbf{$\Delta$Tok$_{15}$} & \textbf{Acc$_{20}$} & \textbf{$\Delta$Acc$_{20}$} & \textbf{Tok$_{20}$} & \textbf{$\Delta$Tok$_{20}$} & \thdr{Math}{Mean} & \thdr{Agentic}{Mean} \\
\midrule
Reference & 67.9 & -- & 9{,}610 & -- & 67.9 & -- & 9{,}610 & -- & 88.0 & 72.6 \\
Student \high   & \win{71.0} & $+3.1$ & 9{,}012 & $-6.2\%$ & 70.7 & $+2.8$ & 9{,}261 & $-3.6\%$ & \win{88.3} & \win{73.1} \\
Student \med    & 68.3 & $+0.4$ & 5{,}160 & $-46.3\%$ & \win{68.8} & $+0.9$ & 5{,}286 & $-45.0\%$ & 83.4 & 72.0 \\
Student \low    & 66.4 & $-1.5$ & \win{3{,}771} & $-60.8\%$ & \win{66.8} & $-1.1$ & 4{,}102 & $-57.3\%$ & 80.3 & 71.2 \\
\bottomrule
\end{tabular}%
}
\caption{\textbf{Multi-teacher OPD on GLM-4.5-Air across 33 benchmarks.} Top four blocks: per-benchmark accuracy at step 15 vs.\ the reference policy; \textbf{(tool)} denotes tool-augmented math and shading marks the best tier. Bottom block: 33-benchmark macro summary at steps 15 and 20.}
\label{tab:aggregate}
\label{tab:full-opd}
\end{table}

\subsection{Transfer from Single-Turn Reasoning to Agent Tasks}
\label{app:rl-transfer}
\Cref{tab:transfer-rl} gives the sparse RL results behind the format-transfer comparison in \cref{sec:pheno-transfer,sec:mech-transfer}. Each entry pairs the task score with mean thinking tokens on a multi-turn benchmark, for the initial policy, the \med and \low policies trained on single-turn reasoning only, and a \med policy trained with agentic data.

Averaged over the nine benchmarks with complete results, the reasoning-only \med policy shortens thinking by $43\%$ (from $7.7$k to $4.4$k tokens) at a score of $79.7$ against $78.7$ initially, and the \low policy shortens it by $61\%$ at $76.9$. The \med policy trained with agentic data reaches a comparable point ($78.0$ at $4.6$k tokens), so the compression obtained from single-turn training is as large as that obtained with in-domain data. The reduction holds on every benchmark, from $26\%$ on $\tau^2$-airline to $83\%$ on \texttt{swe-compass-cu}; the largest score loss is on \texttt{swe-compass-cd} at \low ($90.0$ to $81.0$).
\begin{table}[!htb]
\centering\scriptsize
\setlength{\tabcolsep}{4.8pt}
\renewcommand{\arraystretch}{1.08}
\begin{tabular}{@{}lcccc@{}}
\toprule
\textbf{Benchmark} & \textbf{Initial Policy ($\pi_{\mathrm{init}}$)} & \textbf{Reasoning-Only \med} & \textbf{Reasoning-Only \low} & \textbf{With Agentic Data (\med)} \\
\midrule
\multicolumn{5}{@{}l}{\textit{\textbf{Multi-Turn Tool-Use \& Terminal Environments}}} \\
$\tau^2$-airline       & 86.8 / 6.8k  & \win{86.0} / 5.0k {\tiny($-26\%$)} & 83.2 / 3.4k {\tiny($-50\%$)} & 85.6 / 4.5k {\tiny($-34\%$)} \\
$\tau^2$-retail        & 90.5 / 3.2k  & \win{91.4} / 2.1k {\tiny($-34\%$)} & 90.7 / 1.9k {\tiny($-41\%$)} & 91.0 / 2.2k {\tiny($-31\%$)} \\
$\tau^2$-telecom       & 98.8 / 5.9k  & \win{99.1} / 3.8k {\tiny($-36\%$)} & 97.4 / 3.0k {\tiny($-49\%$)} & 99.0 / 3.7k {\tiny($-37\%$)} \\
Terminal-Bench 2.1     & 64.5 / 27.5k & 62.5 / 18.0k {\tiny($-35\%$)}      & --                           & --                           \\
\midrule
\multicolumn{5}{@{}l}{\textit{\textbf{Multi-Turn Software-Engineering \& Coding-Agent Suites}}} \\
\texttt{swe-compass-cd} & 90.0 / 23.0k & 90.0 / 12.2k {\tiny($-47\%$)}      & 81.0 / 10.1k {\tiny($-56\%$)} & \win{91.0} / 15.2k {\tiny($-34\%$)} \\
\texttt{swe-compass-cu} & 59.3 / 3.5k  & \win{63.5} / 1.3k {\tiny($-63\%$)} & 59.3 / 0.6k {\tiny($-83\%$)}  & 56.3 / 1.5k {\tiny($-57\%$)} \\
\texttt{pev-eval}       & 86.3 / 8.5k  & \win{88.3} / 5.0k {\tiny($-41\%$)} & 88.0 / 2.8k {\tiny($-67\%$)}  & 86.1 / 4.7k {\tiny($-45\%$)} \\
\texttt{pre-eval}       & 87.2 / 9.2k  & \win{88.6} / 5.4k {\tiny($-41\%$)} & 87.7 / 3.2k {\tiny($-65\%$)}  & 86.5 / 4.8k {\tiny($-48\%$)} \\
\texttt{deploy}         & 67.0 / 2.2k  & \win{69.0} / 1.6k {\tiny($-27\%$)} & 65.0 / 0.9k {\tiny($-59\%$)}  & \win{69.0} / 1.6k {\tiny($-27\%$)} \\
\texttt{understand}     & 42.8 / 6.8k  & \win{41.4} / 3.3k {\tiny($-51\%$)} & 40.1 / 1.4k {\tiny($-79\%$)}  & 37.3 / 3.0k {\tiny($-56\%$)} \\
\midrule
\textbf{9-Task Macro Mean} & 78.7 / 7.7k & \win{79.7} / 4.4k {\tiny($-43\%$)} & 76.9 / 3.0k {\tiny($-61\%$)} & 78.0 / 4.6k {\tiny($-40\%$)} \\
\bottomrule
\end{tabular}
\caption{\textbf{Zero-shot transfer of single-turn sparse RL compression to multi-turn agentic benchmarks} (accuracy / mean thinking tokens and $\Delta\bar{L}$ vs.\ $\pi_{\mathrm{init}}$). Trained solely on single-turn reasoning, \med and \low cut agentic tokens by $26\%\text{--}83\%$ while matching in-domain agentic RL (\med, rightmost). The macro mean covers the nine tasks other than Terminal-Bench 2.1.}
\label{tab:transfer-rl}
\end{table}

\subsection{Prompt-Guided RL across Task Categories}
\label{app:recipe-scores}
\Cref{tab:d-export,tab:g-export} break down the Recipe~II results of \cref{sec:prompt-opd} by task category over the 24 benchmarks shared by the two evaluation suites. For DeepSeek-V4-Flash, the comparison includes the reference, the single-tier \med and \low teachers, the prompt-guided policy, and the multi-teacher OPD student. For GLM-4.5-Air, it includes the reference, an earlier family of effort checkpoints, and the prompt-guided policy.

On DeepSeek-V4-Flash, the prompt-guided policy and the OPD student are close in total score at every tier ($66.76$/$66.12$/$63.89$ vs.\ $67.04$/$66.38$/$64.36$), with complementary strengths: the prompt-guided policy is higher on reasoning ($65.25$ vs.\ $63.76$ at \high) and the OPD student is higher on code ($62.29$ vs.\ $58.71$). The code difference comes from Terminal-Bench~2.1: without it, the totals are $67.56$/$66.91$/$65.25$ for the prompt-guided policy and $67.14$/$66.54$/$64.59$ for the OPD student. Agent scores stay close to the reference at all tiers ($70.72$, $71.15$, and $70.14$ vs.\ $71.09$). On GLM-4.5-Air, the prompt-guided \high tier stays within $0.4$ points of the reference in total ($61.47$ vs.\ $61.85$); it gains on general, reasoning, and code benchmarks and loses on agent benchmarks ($60.96$ vs.\ $63.87$). Across tiers, the totals are $61.47$/$60.72$/$59.31$, the agent means $60.96$/$60.24$/$59.10$, and the code means $55.63$/$54.28$/$55.06$, above the reference's $53.93$ at every tier.
% Generated by tables/generate_export_tables.py; do not edit numbers manually.
\begin{table}[!htb]
\centering\footnotesize
\begin{tabular}{@{}lrrrrr@{}}
\toprule
Key & Total & General (4) & Reasoning (6) & Code (5) & Agent (9) \\
\midrule
D0 & 67.80 & 70.81 & 64.40 & 63.54 & 71.09 \\
D1 & 66.07 & 69.92 & 62.01 & 60.34 & 70.25 \\
D2 & 62.37 & 67.28 & 52.90 & 58.36 & 68.72 \\
D3 & 66.76 & 70.16 & 65.25 & 58.71 & 70.72 \\
D4 & 66.12 & 70.14 & 62.22 & 58.53 & 71.15 \\
D5 & 63.89 & 70.03 & 58.12 & 54.66 & 70.14 \\
D6 & 67.04 & 68.92 & 63.76 & 62.29 & 71.04 \\
D7 & 66.38 & 69.36 & 61.35 & 61.61 & 71.07 \\
D8 & 64.36 & 67.45 & 57.34 & 58.76 & 70.77 \\
\bottomrule
\end{tabular}
\caption{DeepSeek-V4-Flash: total and category scores over the 24 benchmarks shared by both evaluations. Parentheses give category sizes. D0: reference; D1--D2: medium/low single-tier teachers; D3--D5: prompt-guided RL (step 20) high/medium/low; D6--D8: multi-teacher OPD high/medium/low. Scores average native benchmark metrics.}
\label{tab:d-export}
\end{table}

% Generated by tables/generate_export_tables.py; do not edit numbers manually.
\begin{table}[!htb]
\centering\footnotesize
\begin{tabular}{@{}lrrrrr@{}}
\toprule
Key & Total & General (4) & Reasoning (6) & Code (5) & Agent (9) \\
\midrule
G0 & 61.85 & 69.26 & 60.50 & 53.93 & 63.87 \\
G1 & 62.70 & 69.77 & 61.45 & 56.13 & 64.03 \\
G2 & 60.75 & 66.30 & 58.17 & 54.74 & 63.34 \\
G3 & 57.46 & 63.64 & 54.15 & 53.19 & 59.31 \\
G4 & 61.47 & 70.38 & 61.16 & 55.63 & 60.96 \\
G5 & 60.72 & 69.42 & 61.00 & 54.28 & 60.24 \\
G6 & 59.31 & 68.22 & 57.22 & 55.06 & 59.10 \\
\bottomrule
\end{tabular}
\caption{GLM-4.5-Air: total and category scores over the 24 benchmarks shared by both evaluations. Parentheses give category sizes. G0: reference; G1--G3: earlier effort checkpoints high/medium/low; G4--G6: prompt-guided RL (step 25) high/medium/low. Scores average native benchmark metrics.}
\label{tab:g-export}
\end{table}

\subsection{Task Scores at Each Thinking Length}
\label{app:recipe-lengths}
The ten general and reasoning benchmarks shared by the two suites have paired measurements of task score and mean thinking tokens. \Cref{tab:d-len,tab:g-len} compare each tier of the prompt-guided policy with its reference, and \cref{tab:d-len-full,tab:g-len-full} extend the comparison to all checkpoints behind the score--length analysis of \cref{fig:score-length}. The last two rows of each table give the mean score and the mean ratio of thinking length to the reference.

The prompt-guided tiers are ordered in length on both models: \high, \med, and \low use $93\%$, $57\%$, and $34\%$ of the reference length on DeepSeek-V4-Flash and $93\%$, $69\%$, and $42\%$ on GLM-4.5-Air. The \high tier scores slightly above the reference on both ($67.22$ vs.\ $66.97$ and $64.85$ vs.\ $64.00$). At matched length on DeepSeek-V4-Flash, the prompt-guided tiers score above the OPD tiers ($67.22$/$65.39$/$62.89$ vs.\ $65.82$/$64.55$/$61.38$ at relative lengths of $0.93$/$0.57$/$0.34$ and $0.92$/$0.55$/$0.33$). The single-tier \low teacher reaches a shorter length ($0.22$) at a lower score ($58.65$).
% Generated by tables/generate_length_tables.py; do not edit numbers manually.
\begin{table}[!htb]
\centering\footnotesize
\setlength{\tabcolsep}{4pt}
\resizebox{\linewidth}{!}{%
\begin{tabular}{@{}lrrrr@{}}
\toprule
Strategy & Reference & \high & \med & \low \\
\midrule
\multicolumn{5}{@{}l}{\emph{General}} \\
AA-LCR & 69.67 / 8.9k & 67.00 / 3.0k & 68.67 / 2.0k & 69.00 / 1.2k \\
LongBench-v2 (128k) & 66.54 / 7.3k & 65.38 / 4.9k & 64.23 / 3.2k & 63.46 / 2.2k \\
IFEval & 88.54 / 1.9k & 89.09 / 2.5k & 87.80 / 1.2k & 87.80 / 536 \\
IFBench & 58.50 / 6.9k & 59.18 / 7.5k & 59.86 / 4.1k & 59.86 / 2.3k \\
\midrule
\multicolumn{5}{@{}l}{\emph{Reasoning}} \\
HLE (judged) & 26.37 / 24.1k & 25.81 / 23.5k & 22.61 / 13.8k & 18.95 / 8.2k \\
GPQA (English) & 88.26 / 11.4k & 90.66 / 11.1k & 89.02 / 6.0k & 87.63 / 3.1k \\
AIME26 (English) & 93.44 / 21.0k & 93.75 / 21.2k & 91.88 / 12.8k & 88.54 / 7.3k \\
HMMT-202602 (English) & 85.70 / 23.9k & 85.04 / 24.2k & 80.59 / 15.4k & 72.92 / 10.2k \\
ARC-AGI-1 & 77.09 / 23.2k & 77.09 / 22.8k & 74.22 / 16.8k & 69.93 / 11.6k \\
ARC-AGI-2 & 15.57 / 43.5k & 19.16 / 42.3k & 14.97 / 30.1k & 10.78 / 21.4k \\
\midrule
Mean score (10) & 66.97 & 67.22 & 65.39 & 62.89 \\
Relative length & 1.00 & 0.93 & 0.57 & 0.34 \\
\bottomrule
\end{tabular}}
\caption{DeepSeek-V4-Flash: paired score and thinking length for the reference and the three tiers of the prompt-guided policy. Each cell reports score / mean thinking tokens per response. Relative length is the mean per-strategy ratio to the reference. Agent and code strategies are omitted because no length statistics are recorded for them.}
\label{tab:d-len}
\end{table}

% Generated by tables/generate_length_tables.py; do not edit numbers manually.
\begin{table}[!htb]
\centering\footnotesize
\setlength{\tabcolsep}{4pt}
\resizebox{\linewidth}{!}{%
\begin{tabular}{@{}lrrrr@{}}
\toprule
Strategy & Reference & \high & \med & \low \\
\midrule
\multicolumn{5}{@{}l}{\emph{General}} \\
AA-LCR & 66.67 / 3.5k & 63.33 / 3.2k & 65.00 / 2.6k & 63.00 / 1.5k \\
LongBench-v2 (128k) & 60.00 / 6.4k & 61.92 / 4.1k & 61.92 / 3.2k & 61.92 / 1.6k \\
MRCR-32k & 81.50 / 3.4k & 80.33 / 4.4k & 81.00 / 2.7k & 77.15 / 1.2k \\
AdvanceIF (export variant) & 50.70 / 1.2k & 54.73 / 1.3k & 52.52 / 855 & 48.35 / 415 \\
IFEval & 87.80 / 1.5k & 90.94 / 1.6k & 88.54 / 951 & 87.43 / 544 \\
IFBench & 62.59 / 4.9k & 65.31 / 5.7k & 62.24 / 3.5k & 60.54 / 2.7k \\
ChineseSimpleQA & 62.65 / 3.0k & 60.05 / 3.3k & 60.45 / 2.4k & 59.05 / 1.1k \\
\midrule
\multicolumn{5}{@{}l}{\emph{Reasoning}} \\
ARC-AGI-1 & 66.11 / 24.6k & 63.25 / 24.8k & 64.68 / 21.6k & 58.47 / 15.2k \\
ARC-AGI-2 & 4.79 / 40.1k & 7.19 / 40.1k & 7.19 / 34.4k & 1.80 / 21.1k \\
HLE (judged) & 25.72 / 23.1k & 24.98 / 19.9k & 23.73 / 12.9k & 19.18 / 5.7k \\
GPQA (English) & 88.64 / 28.5k & 93.18 / 17.8k & 93.56 / 10.1k & 92.30 / 4.2k \\
AIME26 (English) & 94.58 / 10.5k & 94.48 / 10.7k & 93.44 / 8.6k & 91.25 / 5.2k \\
HMMT-202602 (English) & 83.14 / 13.9k & 83.90 / 14.3k & 83.43 / 12.2k & 80.30 / 7.8k \\
\midrule
Mean score (13) & 64.22 & 64.89 & 64.44 & 61.60 \\
Relative length & 1.00 & 0.98 & 0.71 & 0.40 \\
\bottomrule
\end{tabular}}
\caption{GLM-4.5-Air: paired score and thinking length for the reference and the three tiers of the prompt-guided policy. Each cell reports score / mean thinking tokens per response. Relative length is the mean per-strategy ratio to the reference. Agent and code strategies are omitted because no length statistics are recorded for them.}
\label{tab:g-len}
\end{table}

% Generated by tables/generate_length_tables.py; do not edit numbers manually.
\begin{table}[!htb]
\centering\scriptsize
\setlength{\tabcolsep}{4pt}
\resizebox{\linewidth}{!}{%
\begin{tabular}{@{}lrrrrrrrrr@{}}
\toprule
Strategy & D0 & D1 & D2 & D3 & D4 & D5 & D6 & D7 & D8 \\
\midrule
\multicolumn{10}{@{}l}{\emph{General}} \\
AA-LCR & 69.67 / 8.9k & 69.00 / 1.7k & 67.33 / 624 & 67.00 / 3.0k & 68.67 / 2.0k & 69.00 / 1.2k & 67.00 / 5.0k & 71.00 / 1.7k & 68.67 / 973 \\
LongBench-v2 (128k) & 66.54 / 7.3k & 65.38 / 2.8k & 62.31 / 763 & 65.38 / 4.9k & 64.23 / 3.2k & 63.46 / 2.2k & 63.08 / 5.5k & 65.38 / 3.1k & 60.00 / 2.0k \\
IFEval & 88.54 / 1.9k & 87.80 / 1.1k & 85.40 / 407 & 89.09 / 2.5k & 87.80 / 1.2k & 87.80 / 536 & 87.43 / 2.2k & 85.95 / 1.0k & 87.06 / 547 \\
IFBench & 58.50 / 6.9k & 57.48 / 4.6k & 54.08 / 2.0k & 59.18 / 7.5k & 59.86 / 4.1k & 59.86 / 2.3k & 58.16 / 7.5k & 55.10 / 4.2k & 54.08 / 2.4k \\
\midrule
\multicolumn{10}{@{}l}{\emph{Reasoning}} \\
HLE (judged) & 26.37 / 24.1k & 22.10 / 13.7k & 15.62 / 4.2k & 25.81 / 23.5k & 22.61 / 13.8k & 18.95 / 8.2k & 25.67 / 22.7k & 22.06 / 13.1k & 17.79 / 6.7k \\
GPQA (English) & 88.26 / 11.4k & 88.76 / 5.3k & 82.32 / 1.5k & 90.66 / 11.1k & 89.02 / 6.0k & 87.63 / 3.1k & 91.54 / 9.9k & 88.64 / 5.1k & 85.61 / 2.4k \\
AIME26 (English) & 93.44 / 21.0k & 91.04 / 13.3k & 81.67 / 5.2k & 93.75 / 21.2k & 91.88 / 12.8k & 88.54 / 7.3k & 92.50 / 20.5k & 91.25 / 12.9k & 88.65 / 7.6k \\
HMMT-202602 (English) & 85.70 / 23.9k & 82.86 / 16.4k & 67.61 / 6.8k & 85.04 / 24.2k & 80.59 / 15.4k & 72.92 / 10.2k & 84.75 / 22.7k & 82.58 / 15.8k & 75.57 / 10.0k \\
ARC-AGI-1 & 77.09 / 23.2k & 73.51 / 16.9k & 63.01 / 8.2k & 77.09 / 22.8k & 74.22 / 16.8k & 69.93 / 11.6k & 76.13 / 22.1k & 72.79 / 16.8k & 68.02 / 11.8k \\
ARC-AGI-2 & 15.57 / 43.5k & 13.77 / 31.6k & 7.19 / 13.0k & 19.16 / 42.3k & 14.97 / 30.1k & 10.78 / 21.4k & 11.98 / 41.5k & 10.78 / 30.5k & 8.38 / 20.4k \\
\midrule
Mean score (10) & 66.97 & 65.17 & 58.65 & 67.22 & 65.39 & 62.89 & 65.82 & 64.55 & 61.38 \\
Relative length & 1.00 & 0.56 & 0.22 & 0.93 & 0.57 & 0.34 & 0.92 & 0.55 & 0.33 \\
\bottomrule
\end{tabular}}
\caption{DeepSeek-V4-Flash: paired score and thinking length for all checkpoints. Column keys are defined in \cref{tab:d-export}. Each cell reports score / mean thinking tokens per response. Relative length is the mean per-strategy ratio to the reference. Agent and code strategies are omitted because no length statistics are recorded for them.}
\label{tab:d-len-full}
\end{table}

% Generated by tables/generate_length_tables.py; do not edit numbers manually.
\begin{table}[!htb]
\centering\scriptsize
\setlength{\tabcolsep}{4pt}
\resizebox{\linewidth}{!}{%
\begin{tabular}{@{}lrrrrrrr@{}}
\toprule
Strategy & G0 & G1 & G2 & G3 & G4 & G5 & G6 \\
\midrule
\multicolumn{8}{@{}l}{\emph{General}} \\
AA-LCR & 66.67 / 3.5k & 66.67 / 3.4k & 60.67 / 1.1k & 56.33 / 613 & 63.33 / 3.2k & 65.00 / 2.6k & 63.00 / 1.5k \\
LongBench-v2 (128k) & 60.00 / 6.4k & 60.00 / 6.6k & 60.77 / 1.8k & 55.00 / 803 & 61.92 / 4.1k & 61.92 / 3.2k & 61.92 / 1.6k \\
MRCR-32k & 81.50 / 3.4k & 82.06 / 2.3k & 78.78 / 283 & 76.65 / 147 & 80.33 / 4.4k & 81.00 / 2.7k & 77.15 / 1.2k \\
AdvanceIF (export variant) & 50.70 / 1.2k & 51.93 / 1.0k & 49.54 / 498 & 46.14 / 159 & 54.73 / 1.3k & 52.52 / 855 & 48.35 / 415 \\
IFEval & 87.80 / 1.5k & 89.83 / 765 & 87.62 / 314 & 87.80 / 143 & 90.94 / 1.6k & 88.54 / 951 & 87.43 / 544 \\
IFBench & 62.59 / 4.9k & 62.59 / 4.0k & 56.12 / 1.8k & 55.44 / 677 & 65.31 / 5.7k & 62.24 / 3.5k & 60.54 / 2.7k \\
ChineseSimpleQA & 62.65 / 3.0k & 63.04 / 3.4k & 60.45 / 588 & 57.81 / 349 & 60.05 / 3.3k & 60.45 / 2.4k & 59.05 / 1.1k \\
\midrule
\multicolumn{8}{@{}l}{\emph{Reasoning}} \\
ARC-AGI-1 & 66.11 / 24.6k & 67.54 / 24.2k & 56.56 / 14.8k & 47.73 / 8.6k & 63.25 / 24.8k & 64.68 / 21.6k & 58.47 / 15.2k \\
ARC-AGI-2 & 4.79 / 40.1k & 8.98 / 39.5k & 3.59 / 21.8k & 2.40 / 11.8k & 7.19 / 40.1k & 7.19 / 34.4k & 1.80 / 21.1k \\
HLE (judged) & 25.72 / 23.1k & 25.16 / 25.6k & 21.13 / 6.6k & 16.50 / 3.0k & 24.98 / 19.9k & 23.73 / 12.9k & 19.18 / 5.7k \\
GPQA (English) & 88.64 / 28.5k & 88.76 / 33.0k & 92.80 / 4.7k & 89.39 / 1.7k & 93.18 / 17.8k & 93.56 / 10.1k & 92.30 / 4.2k \\
AIME26 (English) & 94.58 / 10.5k & 94.27 / 10.3k & 93.02 / 6.5k & 89.79 / 3.9k & 94.48 / 10.7k & 93.44 / 8.6k & 91.25 / 5.2k \\
HMMT-202602 (English) & 83.14 / 13.9k & 84.00 / 14.0k & 81.91 / 9.0k & 79.07 / 5.7k & 83.90 / 14.3k & 83.43 / 12.2k & 80.30 / 7.8k \\
\midrule
Mean score (13) & 64.22 & 64.99 & 61.77 & 58.47 & 64.89 & 64.44 & 61.60 \\
Relative length & 1.00 & 0.94 & 0.36 & 0.19 & 0.98 & 0.71 & 0.40 \\
\bottomrule
\end{tabular}}
\caption{GLM-4.5-Air: paired score and thinking length for all checkpoints. Column keys are defined in \cref{tab:g-export}. Each cell reports score / mean thinking tokens per response. Relative length is the mean per-strategy ratio to the reference. Agent and code strategies are omitted because no length statistics are recorded for them.}
\label{tab:g-len-full}
\end{table}

\subsection{Benchmark-Level Scores by Checkpoint}
\label{app:checkpoint-scores}
The following tables give the benchmark-level scores behind the category means of \cref{app:recipe-scores}, for GLM-4.5-Air and then DeepSeek-V4-Flash, with one table per category. Column keys follow \cref{tab:g-export,tab:d-export}.

These scores locate the category-level differences on individual benchmarks. On DeepSeek-V4-Flash, the code deficit of the prompt-guided policy comes mostly from Terminal-Bench~2.1 ($48.31$/$47.87$/$32.81$ against $64.94$/$62.70$/$59.10$ for the OPD student). On GLM-4.5-Air, the agent deficit comes from two VitaBench domains, cross-domain and OTA, where all three prompt-guided tiers score between $8$ and $12$ against $17.84$ and $26.25$ for the reference; on \texttt{swe-compass-cd} the same tiers improve on the reference ($81.5$/$80.5$/$83.0$ vs.\ $78.5$).
% Generated by tables/generate_export_tables.py; all exported cells retained.
\begin{table}[!htb]
\centering\scriptsize
\setlength{\tabcolsep}{3pt}
\resizebox{\linewidth}{!}{%
\begin{tabular}{@{}lrrrrrrr@{}}
\toprule
Strategy & G0 & G1 & G2 & G3 & G4 & G5 & G6 \\
\midrule
AA-LCR & 66.67 & 66.67 & 60.67 & 56.33 & 63.33 & 65.00 & 63.00 \\
LongBench-v2 (128k) & 60.00 & 60.00 & 60.77 & 55.00 & 61.92 & 61.92 & 61.92 \\
IFEval & 87.80 & 89.83 & 87.62 & 87.80 & 90.94 & 88.54 & 87.43 \\
IFBench & 62.59 & 62.59 & 56.12 & 55.44 & 65.31 & 62.24 & 60.54 \\
\bottomrule
\end{tabular}}
\caption{GLM-4.5-Air: general strategy scores. Column keys are defined in \cref{tab:g-export}.}
\label{tab:g-general-details}
\end{table}

\begin{table}[!htb]
\centering\scriptsize
\setlength{\tabcolsep}{3pt}
\resizebox{\linewidth}{!}{%
\begin{tabular}{@{}lrrrrrrr@{}}
\toprule
Strategy & G0 & G1 & G2 & G3 & G4 & G5 & G6 \\
\midrule
ARC-AGI-1 & 66.11 & 67.54 & 56.56 & 47.73 & 63.25 & 64.68 & 58.47 \\
ARC-AGI-2 & 4.79 & 8.98 & 3.59 & 2.40 & 7.19 & 7.19 & 1.80 \\
HLE (judged) & 25.72 & 25.16 & 21.13 & 16.50 & 24.98 & 23.73 & 19.18 \\
GPQA (English) & 88.64 & 88.76 & 92.80 & 89.39 & 93.18 & 93.56 & 92.30 \\
AIME26 (English) & 94.58 & 94.27 & 93.02 & 89.79 & 94.48 & 93.44 & 91.25 \\
HMMT-202602 (English) & 83.14 & 84.00 & 81.91 & 79.07 & 83.90 & 83.43 & 80.30 \\
\bottomrule
\end{tabular}}
\caption{GLM-4.5-Air: reasoning strategy scores. Column keys are defined in \cref{tab:g-export}.}
\label{tab:g-reasoning-details}
\end{table}

\begin{table}[!htb]
\centering\scriptsize
\setlength{\tabcolsep}{3pt}
\resizebox{\linewidth}{!}{%
\begin{tabular}{@{}lrrrrrrr@{}}
\toprule
Strategy & G0 & G1 & G2 & G3 & G4 & G5 & G6 \\
\midrule
Terminal-Bench 2.1 & 51.69 & 53.93 & 53.26 & 50.11 & 53.93 & 50.11 & 52.58 \\
swe-compass-cu (CC, 0629) & 60.29 & 62.75 & 61.80 & 58.41 & 62.39 & 61.98 & 61.90 \\
swe-compass-cd (CC) & 78.50 & 78.50 & 74.50 & 78.00 & 81.50 & 80.50 & 83.00 \\
deploy-full-200 (CC-v3) & 54.50 & 57.00 & 59.50 & 56.50 & 55.00 & 54.50 & 53.50 \\
SciCode & 24.65 & 28.47 & 24.65 & 22.92 & 25.35 & 24.31 & 24.31 \\
\bottomrule
\end{tabular}}
\caption{GLM-4.5-Air: code strategy scores. Column keys are defined in \cref{tab:g-export}.}
\label{tab:g-code-details}
\end{table}

\begin{table}[!htb]
\centering\scriptsize
\setlength{\tabcolsep}{3pt}
\resizebox{\linewidth}{!}{%
\begin{tabular}{@{}lrrrrrrr@{}}
\toprule
Strategy & G0 & G1 & G2 & G3 & G4 & G5 & G6 \\
\midrule
pev-eval (OC) & 83.92 & 83.41 & 84.52 & 81.05 & 85.15 & 84.58 & 85.16 \\
pev-eval (CC) & 84.31 & 85.50 & 82.68 & 79.95 & 82.87 & 82.46 & 83.42 \\
tau2-retail (20260319) & 90.00 & 87.72 & 87.37 & 84.39 & 89.82 & 88.77 & 86.14 \\
tau2-airline (20260303) & 83.60 & 83.20 & 83.60 & 76.40 & 83.20 & 78.40 & 78.00 \\
VitaBench cross-domain & 17.84 & 19.50 & 16.00 & 12.00 & 9.25 & 11.75 & 10.00 \\
VitaBench delivery & 50.50 & 51.00 & 51.50 & 44.50 & 45.75 & 41.50 & 41.25 \\
VitaBench instore & 40.85 & 40.25 & 39.25 & 36.50 & 45.75 & 46.25 & 39.50 \\
VitaBench OTA & 26.25 & 27.25 & 26.50 & 21.25 & 8.25 & 10.00 & 10.00 \\
tau2-telecom (20260319) & 97.54 & 98.42 & 98.60 & 97.72 & 98.60 & 98.42 & 98.42 \\
\bottomrule
\end{tabular}}
\caption{GLM-4.5-Air: agent strategy scores. Column keys are defined in \cref{tab:g-export}.}
\label{tab:g-agent-details}
\end{table}

% Generated by tables/generate_export_tables.py; all exported cells retained.
\begin{table}[htbp]
\centering\scriptsize
\setlength{\tabcolsep}{3pt}
\resizebox{\linewidth}{!}{%
\begin{tabular}{@{}lrrrrrrrrr@{}}
\toprule
Strategy & D0 & D1 & D2 & D3 & D4 & D5 & D6 & D7 & D8 \\
\midrule
AA-LCR & 69.67 & 69.00 & 67.33 & 67.00 & 68.67 & 69.00 & 67.00 & 71.00 & 68.67 \\
LongBench-v2 (128k) & 66.54 & 65.38 & 62.31 & 65.38 & 64.23 & 63.46 & 63.08 & 65.38 & 60.00 \\
IFEval & 88.54 & 87.80 & 85.40 & 89.09 & 87.80 & 87.80 & 87.43 & 85.95 & 87.06 \\
IFBench & 58.50 & 57.48 & 54.08 & 59.18 & 59.86 & 59.86 & 58.16 & 55.10 & 54.08 \\
\bottomrule
\end{tabular}}
\caption{DeepSeek-V4-Flash: general strategy scores. Column keys are defined in \cref{tab:d-export}.}
\label{tab:d-general-details}
\end{table}

\begin{table}[!htb]
\centering\scriptsize
\setlength{\tabcolsep}{3pt}
\resizebox{\linewidth}{!}{%
\begin{tabular}{@{}lrrrrrrrrr@{}}
\toprule
Strategy & D0 & D1 & D2 & D3 & D4 & D5 & D6 & D7 & D8 \\
\midrule
HLE (judged) & 26.37 & 22.10 & 15.62 & 25.81 & 22.61 & 18.95 & 25.67 & 22.06 & 17.79 \\
GPQA (English) & 88.26 & 88.76 & 82.32 & 90.66 & 89.02 & 87.63 & 91.54 & 88.64 & 85.61 \\
AIME26 (English) & 93.44 & 91.04 & 81.67 & 93.75 & 91.88 & 88.54 & 92.50 & 91.25 & 88.65 \\
HMMT-202602 (English) & 85.70 & 82.86 & 67.61 & 85.04 & 80.59 & 72.92 & 84.75 & 82.58 & 75.57 \\
ARC-AGI-1 & 77.09 & 73.51 & 63.01 & 77.09 & 74.22 & 69.93 & 76.13 & 72.79 & 68.02 \\
ARC-AGI-2 & 15.57 & 13.77 & 7.19 & 19.16 & 14.97 & 10.78 & 11.98 & 10.78 & 8.38 \\
\bottomrule
\end{tabular}}
\caption{DeepSeek-V4-Flash: reasoning strategy scores. Column keys are defined in \cref{tab:d-export}.}
\label{tab:d-reasoning-details}
\end{table}

\begin{table}[!htb]
\centering\scriptsize
\setlength{\tabcolsep}{3pt}
\resizebox{\linewidth}{!}{%
\begin{tabular}{@{}lrrrrrrrrr@{}}
\toprule
Strategy & D0 & D1 & D2 & D3 & D4 & D5 & D6 & D7 & D8 \\
\midrule
SciCode & 31.94 & 30.21 & 29.17 & 29.51 & 29.86 & 29.17 & 31.60 & 31.60 & 29.86 \\
swe-compass-cd (CC) & 92.50 & 93.00 & 92.00 & 92.00 & 89.50 & 90.00 & 94.50 & 92.00 & 90.50 \\
swe-compass-cu (CC, 0629) & 58.10 & 57.81 & 53.92 & 57.74 & 56.91 & 55.33 & 58.43 & 58.25 & 55.32 \\
Terminal-Bench 2.1 & 63.15 & 54.16 & 55.73 & 48.31 & 47.87 & 32.81 & 64.94 & 62.70 & 59.10 \\
deploy-full-200 (CC-v3) & 72.00 & 66.50 & 61.00 & 66.00 & 68.50 & 66.00 & 62.00 & 63.50 & 59.00 \\
\bottomrule
\end{tabular}}
\caption{DeepSeek-V4-Flash: code strategy scores. Column keys are defined in \cref{tab:d-export}.}
\label{tab:d-code-details}
\end{table}

\begin{table}[!htb]
\centering\scriptsize
\setlength{\tabcolsep}{3pt}
\resizebox{\linewidth}{!}{%
\begin{tabular}{@{}lrrrrrrrrr@{}}
\toprule
Strategy & D0 & D1 & D2 & D3 & D4 & D5 & D6 & D7 & D8 \\
\midrule
pev-eval (CC) & 86.99 & 83.84 & 83.83 & 85.42 & 84.64 & 85.69 & 87.60 & 86.17 & 86.37 \\
pev-eval (OC) & 75.46 & 74.20 & 84.78 & 76.92 & 75.03 & 77.08 & 87.38 & 87.14 & 85.33 \\
tau2-retail (20260319) & 90.88 & 91.40 & 90.00 & 90.70 & 92.63 & 91.58 & 91.75 & 90.53 & 90.35 \\
tau2-airline (20260303) & 85.20 & 87.60 & 82.80 & 83.20 & 86.80 & 84.80 & 84.40 & 84.00 & 86.00 \\
VitaBench cross-domain & 30.75 & 28.00 & 26.50 & 32.00 & 32.00 & 31.00 & 29.50 & 27.00 & 23.75 \\
VitaBench delivery & 65.25 & 62.75 & 57.75 & 60.00 & 64.00 & 62.00 & 59.00 & 62.00 & 58.25 \\
VitaBench instore & 60.75 & 58.25 & 54.00 & 62.00 & 60.75 & 60.75 & 58.25 & 59.00 & 60.75 \\
VitaBench OTA & 44.75 & 47.25 & 40.60 & 47.50 & 47.00 & 42.25 & 42.00 & 44.50 & 47.75 \\
tau2-telecom (20260319) & 99.82 & 98.95 & 98.25 & 98.77 & 97.54 & 96.14 & 99.47 & 99.30 & 98.42 \\
\bottomrule
\end{tabular}}
\caption{DeepSeek-V4-Flash: agent strategy scores. Column keys are defined in \cref{tab:d-export}.}
\label{tab:d-agent-details}
\end{table}

\section{Proofs for \Cref{sec:mechanism}}
\label{app:proofs}

Assume responses have a finite cap and distributions share the support needed for finite KL. The same inequalities hold in the extended sense when a KL term is infinite.

\subsection{Proof of \Cref{prop:gspo-invariance}}
Let $b_j=g_j-\bar g$. When verifier outcomes are identical, $r_j-\bar r=-\lambda b_j$ and $\operatorname{std}(\{r_j\})=\lambda\sigma_g$ for $\lambda\ge0$. Substitution into the stabilized advantage yields $-\lambda b_j/(\lambda\sigma_g+\epsilon)$. If $\sigma_g=0$, then $b_j=0$ for every rollout, so the advantage is zero. For $\sigma_g>0$ and $\lambda>0$, divide numerator and denominator by $\lambda\sigma_g$ to obtain
\begin{equation}
-\frac{b_j/\sigma_g}{1+\epsilon/(\lambda\sigma_g)}
\longrightarrow -\frac{b_j}{\sigma_g}.
\label{eq:proof-gspo-limit}
\end{equation}
For $\lambda=0$, the advantage is zero under the identical-verifier assumption.

\subsection{Proof of \Cref{prop:opd-stopping}}
Write $a=p_i(e)$ and $b=q_i(e)$. The log-sum inequality applied to all $u\ne e$ gives
\begin{equation}
\KL(p_i\|q_i)=a\log\frac{a}{b}+\sum_{u\ne e}p_i(u)\log\frac{p_i(u)}{q_i(u)}
\ge a\log\frac{a}{b}+(1-a)\log\frac{1-a}{1-b}=\dB(a\|b).
\label{eq:proof-logsum}
\end{equation}
By the chain rule, \cref{eq:opd} is the expected sum of next-token KL divergences over student-visited states. Every exact KL term is nonnegative. Drop terms after $\min(\Lth(y)+1,|y|)$ and apply the pointwise inequality to the retained terms to obtain the stated bound. A single-token stopping marker is an explicit assumption; a multi-token marker requires a state representation that tracks its completion.

\end{document}